\documentclass[journal]{IEEEtran}
\usepackage{cite}
\usepackage{amsmath,amssymb,amsfonts}
\usepackage{algorithmic}
\usepackage{graphicx}
\usepackage{textcomp}
\usepackage{times}
\usepackage{latexsym}
\usepackage{float}
\usepackage{xcolor}
\usepackage{caption}
\usepackage{soul}
\usepackage[maxfloats=512]{morefloats}
\usepackage[T1]{fontenc}

\usepackage[utf8]{inputenc}

\usepackage{microtype}

\usepackage{inconsolata}

\usepackage{hyperref}       
\usepackage{url}            
\usepackage{booktabs}       
\usepackage{amsfonts}       
\usepackage{nicefrac}       
\usepackage{microtype}      
\usepackage{booktabs}       
\usepackage{float}          
\usepackage{cleveref} 
\extrafloats{1000}
\usepackage[dvipsnames]{xcolor}
\newcommand{\best}[1]{\textcolor{ForestGreen}{#1}}
\usepackage[normalem]{ulem} 

\def\BibTeX{{\rm B\kern-.05em{\sc i\kern-.025em b}\kern-.08em
    T\kern-.1667em\lower.7ex\hbox{E}\kern-.125emX}}
\begin{document}

\title{A Large-Scale Evaluation of Role-Enhanced Zero-Shot Chain-of-Thought Prompting}

\author{Ibrahim Khan,
        Thai Van Nguyen,
        Yi Xia,
        Pratch Suntichaikul,
        Mustafa Can Gursesli,
        Daniel Fernandez Galeote,
        Juho Hamari,
        and~Ruck Thawonmas%
\thanks{Ibrahim Khan and Ruck Thawonmas are with the College of Information Science and Technology, Ritsumeikan University, Japan}
\thanks{Thai Van Nguyen and Yi Xia are with the Graduate School of Information Science and Technology, Ritsumeikan University, Japan.}
\thanks{Pratch Suntichaikul is an Independent Researcher, and an alumnus of the Graduate School of Information Science and Technology, Ritsumeikan University, Japan.}
\thanks{Mustafa Can Gursesli, Daniel Fernandez Galeote, and Juho Hamari are with the Research Centre of Gameful Realities, Gamification Group, Tampere University, Finland.}}

\maketitle

\begin{abstract}
This paper proposes Role-Enhanced Zero-Shot Chain-of-Thought (RE-ZS-CoT), a prompting framework that systematically integrates domain-specific expert roles to enhance reasoning in large language models (LLMs) without task-specific examples. We assess two versions of our approach. In the first version, static role selection, we evaluate eight predefined expert roles across arithmetic, commonsense, and reading comprehension benchmarks to determine their impact on reasoning performance. In the second version, dynamic role selection, each model autonomously selects the single most suitable role for each benchmark, enabling adaptive task-role alignment without manual specification. Across 20 LLMs and seven benchmarks, RE-ZS-CoT frequently surpasses zero-shot and zero-shot-CoT baselines, achieving performance improvements of up to 44 percentage points. The results reveal a strong correspondence between static and dynamic versions; the benchmarks where static roles yielded improvements were almost always those where dynamic roles also outperformed baselines, though static selection achieved peak performance. Ablation studies further show that unconventional roles degrade performance, underscoring the necessity of task-role alignment. The framework’s scalability is validated across model sizes (2b–14b parameters), with smaller models like Granite3.3-8b achieving remarkable proficiency in arithmetic reasoning. RE-ZS-CoT advances zero-shot prompting by unifying role specialization with chain-of-thought reasoning, offering a practical, resource-efficient solution for improving LLM capabilities across diverse domains.
\end{abstract}

\begin{figure*}
  \centering
  \includegraphics[width=\textwidth]{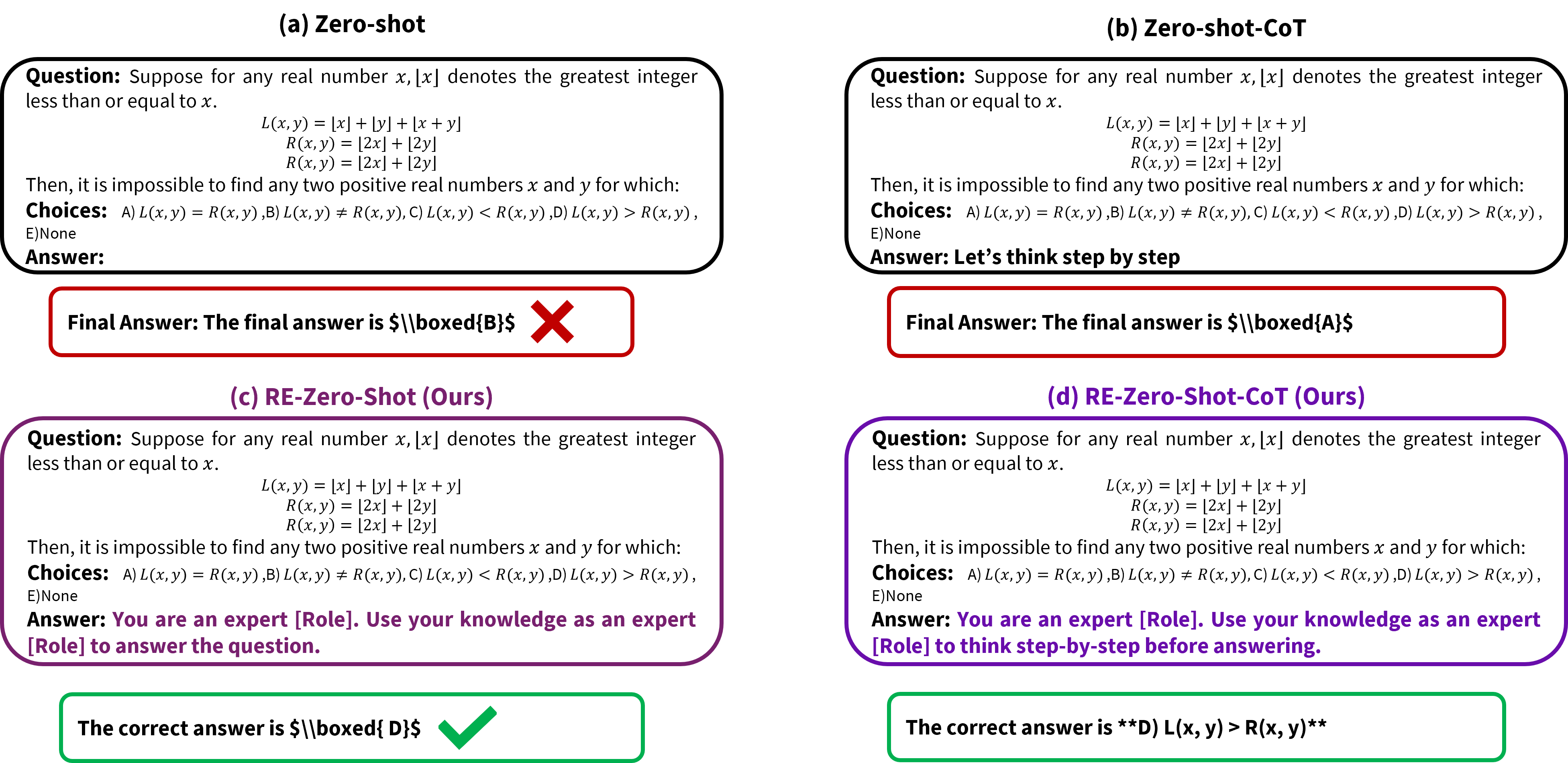}

  \caption{Examples of prompts and corresponding responses generated by Gemma3:12b model on the AQuA-RAT benchmark comparing zero-shot and zero-shot-CoT baselines to our RE-ZS-CoT and RE-ZS. The reasoning of the model is omitted for readability.}
  \label{examplep}
\end{figure*}
\section{Introduction}

Large language models (LLMs) exhibit enhanced reasoning capabilities when directed by structured prompting techniques, such as chain-of-thought (CoT) \cite{wei2022chain} and zero-shot-CoT \cite{kojima2022large}. However, these methodologies frequently demonstrate inconsistent performance across various tasks, particularly in zero-shot scenarios where task-specific examples are unavailable. Recent studies explore various strategies to enhance reasoning in large language models, including decoding techniques to elicit CoT reasoning without explicit prompts \cite{wang2024chain}, structured prompting methods like Plan-and-Solve to improve zero-shot reasoning \cite{zhou2023plan}, and frameworks such as Graph of Thoughts that model LLM outputs as graphs for complex problem-solving \cite{besta2024graph}. Additionally, LLMs demonstrate improved reasoning by taking on a persona, which often leads to greater accuracy, though not consistently \cite{salewski2023context}, \cite{kong2024better}. However, evaluations of role-enhanced prompting, particularly regarding its scalability, adaptability, and effects across reasoning domains, remain unexplored. \par

We present Role-Enhanced Zero-Shot Chain-of-Thought (RE-ZS-CoT), a prompting framework that integrates expert role specification into single-turn, zero-shot reasoning without requiring task examples (Fig. \ref{examplep}). Unlike previous work relying on fixed templates (\cite{salewski2023context}, \cite{kong2024better}, \cite{yuan2024instance}), RE-ZS-CoT fills the above-stated gap by evaluating two versions. Version 1 assesses eight roles across three prompt variants (pre-question, post-question, split-phrase; Fig.\ref{varis}) and uses self-alignment, leveraging recent advances in aligning LLM behavior \cite{sun2023principle}; \cite{wang2022self}; \cite{madaan2023self}. Version 2 shifts to an adaptive framework where each LLM independently selects the most suitable role for each benchmark, enabling self-optimization of task-role alignment without manual adjustment, using the same three prompt variants. \par

In version 1 (static self-alignment), we identify optimal roles for open-weight models under computational constraints. For OpenAI models, we test all eight roles on GPT-4o-mini, while larger models (GPT-4o, GPT-4.1) use only the best-performing role due to cost. Open-weight models (2–3b, 7–9b, 12–14b) use quantized instruct-tuned versions (q4-K-M). We evaluate 20 LLMs across seven benchmarks spanning arithmetic, commonsense, and reading comprehension. RE-ZS-CoT yields 12–44\% higher accuracy than zero-shot and zero-shot-CoT baselines, with the largest gains in arithmetic tasks (AQuA-RAT, MATH). Role effects vary: domain-specific roles (e.g., mathematician) outperform general ones in arithmetic structured tasks, consistent with findings from a previous study \cite{kong2024better}. \par

The contributions of this paper are: (1) We introduce RE-ZS-CoT, the first framework that unifies role-enhancement with zero-shot-CoT reasoning, demonstrating broad applicability across seven benchmarks and 20 LLMs. (2) We show that the efficacy of role specialization is task-dependent, with math reasoning benefiting the most from domain-aligned roles. (3) We provide actionable insights for prompt engineering, highlighting the superiority of pre-question prompts and adaptive role selection for open-weight models. (4) Our ablation studies reveal novel findings regarding the impact of non-traditional roles and lexical variations, advancing the design of context-aware prompting strategies. \par

\section{Related Work}
\subsection{Chain-of-Thought Prompting for Reasoning.}
CoT prompting enhances complex reasoning by eliciting intermediate steps \cite{wei2022chain}. \cite{kojima2022large} showed zero-shot-CoT via simple prompts like "Let’s think step by step," but such prompts remain sensitive to phrasing, producing unstable outputs. Techniques like self-consistency decoding \cite{wang2022self2} and instruction tuning \cite{sanh2021multitask} improve the reliability of CoT and zero-shot prompting, respectively, but rely heavily on generic or example-based prompts. Smaller models (<100B parameters) particularly struggle without fine-tuning \cite{kim2023cot}. Syntactic variations were explored by \cite{reynolds2021prompt}, but without domain-specific reasoning. Recent work emphasizes instance-adaptive prompts \cite{yuan2024instance}, variability among equivalent prompts \cite{cao2024worst}, and reasoning via decoding alone \cite{wang2024chain}, highlighting the need for methods like RE-ZS-CoT to improve zero-shot reliability without extensive prompt engineering.

Recent advancements in reasoning architectures have shifted the paradigm toward models that natively generate reasoning traces via large-scale reinforcement learning (RL), often referred to as test-time compute or System 2 thinking models \cite{jaech2024openai,guo2025deepseek}. Unlike standard CoT prompting, which relies on explicit user instructions to elicit step-by-step logic, these native reasoning models automatically allocate additional computational resources at inference time to plan, search, and self-correct. While these approaches achieve state-of-the-art performance on complex benchmarks, they introduce substantial latency and inference-time overhead. In contrast, RE-ZS-CoT explores the orthogonal direction of maximizing the reasoning efficiency of standard, non-RL-tuned language models by leveraging domain-specific personas.

\subsection{Role-Based Prompting}
Role-enhanced prompting improves LLM reasoning by biasing models toward domain expertise. \cite{salewski2023context} show that instructing models to "act as a domain expert" boosts accuracy, and \cite{kong2024better} demonstrate that role-playing enhances zero-shot reasoning. However, prior methods rely on curated examples \cite{shanahan2023role}, narrow domains \cite{kong2024better}, or manual prompt templates \cite{zheng2024when}. These results create inconsistencies and reliance on a rigid template, which is not always effective. Moreover, the research on role prompting has not been conducted on a large scale, particularly in the zero-shot-CoT domain. Recent work has systematically evaluated social roles in system prompts, finding that interpersonal roles such as "friend" or "teacher" outperform generic "helpful assistant" across diverse tasks, although optimal role prediction remains challenging \cite{zheng2023helpful}. Others explore role-playing for domain adaptation, using unique role prompts per domain alongside self-distillation to prevent catastrophic forgetting during multi-domain fine-tuning \cite{wang2024role}. Interpretable steering via sparse autoencoders extracts role-specific features for fine-grained control over reasoning behavior \cite{wang2025improving}.

In mathematical reasoning, role-play prompts such as "helpful mathematician," combined with zero-shot-CoT, were tested but showed no improvements and sometimes degradation due to role-task mismatch and over-prompting \cite{han2024rethinking}. These works, often limited to a few datasets and models without testing role placement variants or dynamic selection, underscore RE-ZS-CoT's advantages: it scales across 20 LLMs and seven benchmarks, incorporates static/dynamic role selection with prompt variants (pre-question, post-question, split), and unifies role specialization with zero-shot CoT in single-turn prompts, achieving 12–44\% higher accuracy.


\begin{figure}
  \centering
  
 \includegraphics[width=\linewidth]{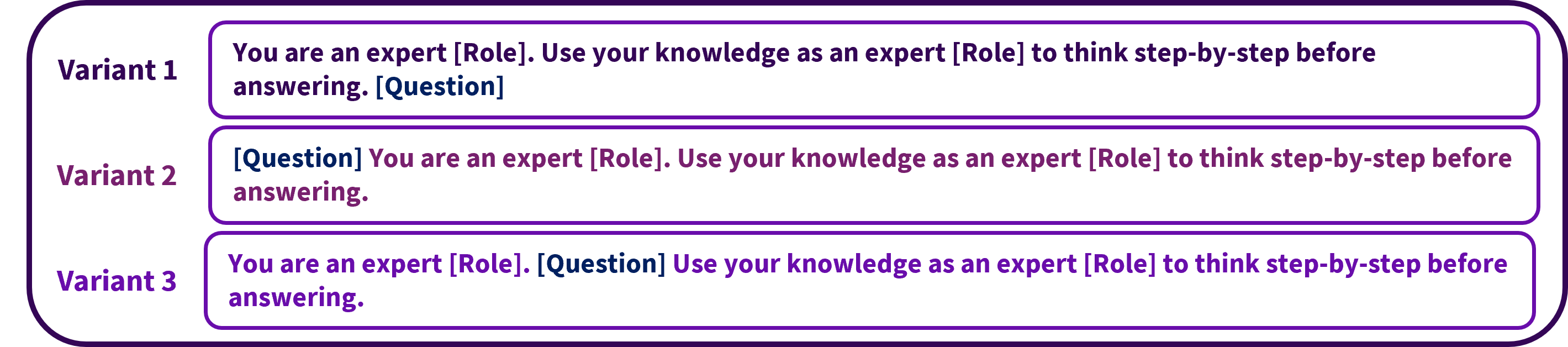}
  \caption{Our approach's pre and post questions, and split-phrase variants with example prompts. Figure \ref{examplep} demonstrates the 2nd variant in action.}
  \label{varis}
\end{figure}
\vspace{-2mm}

\begin{figure*}
  \centering
  \includegraphics[width=\textwidth]{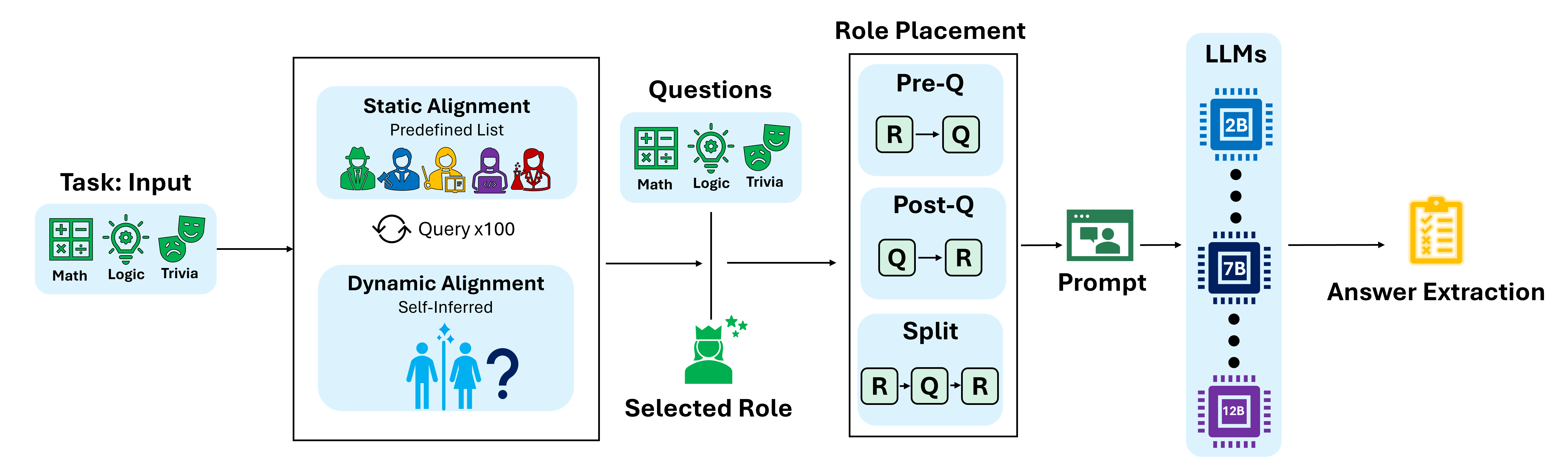}
  \caption{RE-ZS-CoT Framework.}
  \label{Ssytem}
\end{figure*}
\section{Role-Enhanced Zero-Shot Chain of Thought}
\label{sec:3}
We introduce RE-ZS-CoT as shown in Figure \ref{Ssytem} , a prompting technique that integrates role-based prompting within a zero-shot-CoT reasoning framework. Building on previous research \cite{kojima2022large,kong2024better,salewski2023context,han2024rethinking}, our approach adheres to a zero-shot methodology for CoT reasoning while incorporating role-enhanced prompts to bolster the reasoning capabilities of LLMs. Distinct from traditional zero-shot-CoT, which relies on a generic template applicable across diverse tasks and often leads to suboptimal performance in specific evaluations, such as those requiring commonsense reasoning, we utilize two versions of our approach.  \par
In the first version, there are eight specialized roles: \textit{Mathematician, Physicist, Physician, Scientist, Literacy Specialist, Engineer, Software Developer, and Polymath}. This targeted role assignment allows us to create prompts that are more finely attuned to the requirements of each task. The second version examines whether a more adaptive, on-the-fly selection can enhance the roles identified in version 1. For this, we implemented a dynamic role selection mechanism. In this experimental setup, we prompted each model to choose the single most suitable role for each specific benchmark. In addition, unlike the two-step prompting approach employed in zero-shot-CoT, where, according to its original form, reasoning is first extracted before generating the final answer, RE-ZS-CoT leverages a single prompt to extract answers from the model’s response. Figure \ref{examplep} shows an example of our approach compared to zero-shot and zero-shot-CoT.

It is important to note that we do not benchmark RE-ZS-CoT against existing role-prompting frameworks. The fundamental integration of role-playing with zero-shot chain-of-thought reasoning is not in itself a novel concept; prior studies have already established that domain-specific personas can improve reasoning \cite{kong2024better,han2024rethinking}. However, existing literature has largely overlooked the critical mechanics of how these roles are selected and where the role phrases are placed within the prompt. Therefore, our approach serves as a rigorous, large-scale evaluation of these underexplored dimensions. We introduce unique role selection techniques, specifically contrasting static, predefined assignments with dynamic, autonomous inference, and systematically investigate the structural impact of role-phrase placement (R→Q, Q→R, and R→Q→R). Because our primary contribution lies in isolating the effects of these specific selection and placement strategies across 20 distinct LLMs, we strictly compare our framework against standard zero-shot and zero-shot-CoT baselines rather than other specialized role-prompting methods.
\subsection{Prompt Variants}
Our prompt consists of: (i) a role prompt R and (ii) the task question Q. We evaluate three placement strategies, as shown in Fig. \ref{varis}: R→Q, Q→R, and the split form R→Q→R. This setup allows us to examine whether the placement of the role phrase affects the performance of LLMs, and whether defining a role beforehand, as done in a previous study \cite{kong2024better}, can enhance the LLM’s role-playing capabilities. Prior studies have shown that role-play phrases can improve factuality and reasoning (\cite{kong2024better}. Our strategy tests whether these improvements persist under strict zero-shot settings at scale, across LLMs of varying sizes and benchmarks.
\vspace{-2mm}
\section{Implementation Details}
 \label{sec:4}
\paragraph{Benchmarks}
We evaluate our approach on seven datasets in three categories: \textit{arithmetic reasoning, commonsense reasoning, and reading comprehension}. These datasets are commonly used in the benchmarking of LLMs. For a detailed description of each dataset, see [Appendix \ref{sec:A}]. \par

For arithmetic reasoning, we consider the following three datasets: AQuA-RAT \cite{Aqua}, and GSM8K \cite{gsm8k}. AQuA-RAT contains algebraic word problems. It is designed to assess a model's ability not only to solve the problem but also to generate a detailed, step-by-step rationale for its answer in natural language. GSM8K contains grade school math word problems that are linguistically diverse. It primarily checks a model's ability to perform multi-step reasoning, as the questions require between two and eight steps using basic arithmetic to arrive at the final answer. \par

For commonsense reasoning, we use OpenBookQA \cite{OpenBookQA2018}, and StrategyQA \cite{strategyqa}. OpenBookQA contains questions that require multi-step reasoning and the use of commonsense knowledge. StrategyQA requires models to infer an implicit multi-hop reasoning to answer questions. \par

For reading comprehension, we use RACE (h\&m) \cite{race} and TriviaQA \cite{triviaqa}. RACE is a reading comprehension dataset containing questions from the examination of middle (RACE-m) and high school (RACE-h) students. TriviaQA is a reading comprehension dataset containing question-answer pairs authored by trivia enthusiasts and independently gathered evidence documents.

\paragraph{Large Language Models}
We conduct experiments utilizing a total of 20 LLMs, ranging from 2b to 14b open-weights models along with large closed-source models from OpenAI. This exploration aims to investigate the impact of our approach on varying parameter sizes, extending from large to small. For the main experiments, we use all 20 models, which are listed as follows: GPT-4.1, GPT-4.1-mini, GPT-4.1-nano, GPT-4o, GPT-4o-mini, Phi4:14b \cite{Phi4}, Qwen2.5:14b \cite{Qwen2.5}, Gemma3:12b \cite{Gemma3}, Mistral-nemo:12b-instruct \cite{Mistralnemo}, Gemma2:9b-instruct \cite{Gemma2}, Granite3.2:8b-instruct \cite{Granite32}, Granite3.3:8b \cite{Granite33}, Llama3.1:8b-instruct \cite{Llama3.1}, Mistral:7b-instruct \cite{jiang2023Mistral7b}, Qwen2.5:7b-instruct \cite{Qwen2.5}, Llama3.2:3b-instruct \cite{meta2024Llama32}, Qwen2.5:3b-instruct \cite{Qwen2.5}, Granite3.2:2b-instruct \cite{Granite32}, Granite3.3:2b \cite{Granite33}, and Gemma2:2b-instruct \cite{Gemma2}. \par

All the open-weights models are used with the latest available versions at the time of writing. We explicitly list the full names of all models [see Appendix \ref{sec:A.2}], including the closed-source OpenAI models, to ensure traceability in case of future updates.

\paragraph{Baselines and Answer Detection}
We compare our approach with standard zero-shot prompting and zero-shot-CoT. This is done to verify the effectiveness of our approach as we measure both zero-shot-CoT and zero-shot versions of our method. For our experiments, we use the same prompt for zero-shot and zero-shot-CoT, with the difference being the added \textit{“Let’s think step by step.”} at the end of the question for zero-shot-CoT; see Fig. \ref{examplep} for an example. In contrast to the two-stage prompting employed in prior research conducted by \cite{kojima2022large}, we utilize a one-stage approach for all prompts, including zero-shot-CoT, to ensure an equitable comparison.  \par

For answer detection in evaluation, drawing inspiration from previous research \cite{kojima2022large}, we examine the model’s output and select the first token that aligns with the dataset’s answer specifications (for instance, a capital letter A–E for multiple-choice questions, the first valid numeral for arithmetic problems. The captured string is subjected to a cleaning process: punctuation, LaTeX wrappers, currency symbols, and thousands separators are removed, and logical terms (True/False) are converted to their corresponding letter equivalents prior to being returned as the prediction. In instances where we fail to identify an explicit cue, such as \textit{“The answer is …,”}, we resort to this reverse scanning method; if no matches are found, we document a failure and classify the item as incorrect. In summary, all tasks adhere to the same procedure: identify the earliest compliant answer token, normalize it, and compare it to the answer label. 

\paragraph{Static Role Self-Alignment}
For the static roles, we asked the open-weights LLMs themselves to select the most suitable roles from a predefined list based on the provided task description. Each LLM was queried 100 times, and we selected the two most frequently chosen roles to use in the experiments. This approach is inspired by previous work \cite{sun2023principle}; \cite{wang2022self}; \cite{madaan2023self}, which leverages LLMs to self-align themselves on specific tasks. We adopted this method for several reasons, besides our financial and computational constraints. Namely, unlike complete pilot runs, which require testing all role-prompt combinations, this strategy uses just a few lightweight meta-prompts per benchmark, making it ideal for resource-constrained settings. As new tasks or roles emerge, the same self-assessment procedure can be reused without manual ontology construction or extensive sampling. This property can offer a notable engineering advantage. Full details are provided in [Appendix~\ref{sec:B}]. \par

For closed-source models, we executed all roles for each task on GPT-4o-mini and subsequently selected the most effective role for the larger and more costly models (GPT-4o, GPT-4.1), considering financial limitations. Furthermore, we implemented all roles for each task on GPT-4.1-mini and nano. We tested all roles on GPT-4o-mini, GPT-4.1-mini, and nano to not only cross-validate the self-alignment approach but also to assess the performance of all roles on all tasks for a broader impact and analysis. \par

\paragraph{Dynamic Role Self-Alignment}
In order to address the possibility that the optimal role for a given task is not in a predefined list, a second version of our framework is introduced that enables models to infer their own task-specific roles autonomously. In this version, each LLM is prompted to determine the single most suitable role for solving a given benchmark based solely on its understanding of the task description. Unlike the static setting, no predefined list of roles is provided. Instead, the model generates the role it deems most appropriate. Same as static self-alignment, we queried each LLM 100 times and selected the role most mentioned by each LLM. \par

This mechanism transforms role selection into a fully adaptive, model-driven process, allowing each LLM to express its internal notion of domain expertise. Similar to the static self-alignment method, this dynamic variant leverages intrinsic meta-reasoning capabilities but extends them to an open-role space, where models infer both the identity and relevance of the role without prior constraints. Full implementation details, including meta-prompt examples and role frequency statistics, are provided in [Appendix~\ref{sec:B}].

\section{Main Results}
 \label{sec:5}
 \subsection{Overall Results}

 \begin{table*}[ht]
    \centering
    \caption{Results of all models on the AQuA-RAT, GSM8K, and OpenbookQA datasets. The results are presented in zero-shot/zero-shot-CoT/Ours (static)/Ours (dynamic) format. \uline{Underlined} scores indicate when our approach outperforms the baselines but is not the best result. Symbol “$\equiv$” indicates that the dynamic role was identical to the static role.}
    \label{Oresults1}
    \resizebox{\textwidth}{!}{
      \begin{tabular}{lcccccccccccccccc}
        \toprule
        & \multicolumn{4}{c}{\textbf{AQuA-RAT}} 
        & \multicolumn{4}{c}{\textbf{GSM8K}} 
        & \multicolumn{4}{c}{\textbf{OpenbookQA}}  \\
        \cmidrule(lr{.5em}){2-5} 
        \cmidrule(lr{.5em}){6-9} 
        \cmidrule(lr{.5em}){10-13} 
        \textbf{Model} 
          & \textbf{0S} & \textbf{0CoT} & \textbf{Ours (S)} & \textbf{Ours (D)}
          & \textbf{0S} & \textbf{0CoT} & \textbf{Ours (S)} & \textbf{Ours (D)}
          & \textbf{0S} & \textbf{0CoT} & \textbf{Ours (S)} & \textbf{Ours (D)} \\
        \midrule
        GPT4.1 
          & 80.31 & 67.72 & \best{82.68} & 68.50
          & 86.81 & 86.96 & 85.75 & $\equiv$
          & 79.40 & 87.20 & \best{94.4} & \uline{93.00}\\
        GPT-4.1-mini 
          & 83.07 & 85.04 & \best{86.22} & 78.35
          & 89.16 & 87.04 & 87.79 & 87.11
          & 89.00 & 96.20 & 96.20 & \best{96.40} \\
        GPT-4.1-nano 
          & 75.98 & 71.65 & \best{77.95} & 70.87
          & 87.87 & 86.73 & 86.66 & 85.37
          & 87.40 & 86.20 & \best{91.40} & \uline{90.00} \\
        GPT-4o 
          & 72.44 & 71.65 & \best{80.31} & $\equiv$
          & 88.63 & 89.08 & 88.48 & $\equiv$
          & 96.60 & 95.60 & 95.00 & 92.60\\
        GPT-4o-mini 
          & 67.32 & 69.29 & \best{78.74} & \uline{73.62}
          & 88.48 & 88.02 & \best{88.78} & $\equiv$
          & 92.60 & 80.20 & 84.80 & 83.60 \\
        Phi4-14b 
          & 70.08 & 68.50 & \best{74.02} & 65.35
          & 87.72 & 87.34 & \best{89.99} & \uline{89.31}
          & 73.20 & 64.60 & 40.00 & 61.20\\
        Qwen2.5-14b 
          & 66.93 & 69.69 & 64.57 & 60.24
          & 86.58 & 87.72 & \uline{88.02} & \best{88.25}
          & 61.40 & 62.60 & 57.00 & 55.40 \\
        Gemma3-12b 
          & 20.87 & 23.23 & \best{67.32} & \uline{58.66}
          & 86.20 & 85.22 & \uline{87.41} & \best{88.02}
          & 85.39 & 32.80 & 51.40 & 49.00 \\
        Mistral-nemo-12b 
          & 59.06 & 61.02 & 59.45 & 59.84
          & 77.26 & 78.85 & \best{80.82} & $\equiv$
          & 41.60 & 63.60 & \uline{75.20} & \best{77.40}\\
        Gemma2-9b 
          & 54.72 & 51.97 & \best{55.51} & 51.57
          & 84.91 & 84.38 & 84.84 & 81.96
          & 88.60 & 47.60 & 36.80 & 33.60 \\
        Granite3.2-8b 
          & 53.54 & 53.94 & \best{60.63} & \uline{59.45}
          & 75.51 & 77.33 & \best{79.38} & 77.10
          & 66.80 & 62.20 & 66.60 & 62.80\\
        Granite3.3-8b 
          & 55.51 & 59.06 & \best{62.60} & \uline{60.63}
          & 75.89 & 75.82 & \best{79.30} & \uline{78.24}
          & 75.80 & 74.20 & 75.40 & 75.20\\
        Llama3.1-8b 
          & 49.21 & 51.57 & \best{60.24} & $\equiv$
          & 79.91 & 80.36 & \best{80.74} & 80.36
          & 52.60 & 59.80 & 51.80 & 55.00\\
        Mistral-7b 
          & 12.60 & 16.14 & \best{20.08} & $\equiv$
          & 34.27 & 42.38 & \best{42.84} & \uline{42.53}
          & 52.20 & 44.60 & 37.60 & 41.60\\
        Qwen2.5-7b 
          & 74.41 & 76.38 & 70.87 & $\equiv$
          & 84.46 & 84.53 & \best{85.97} & \uline{85.29}
          & 31.80 & 50.00 & \uline{62.00} & \best{67.00} \\
        Llama3.2-3b 
          & 55.91 & 51.57 & 51.97 & 48.21
          & 71.57 & 73.62 & 73.31 & 73.16
          & 30.20 & 58.60 & \uline{60.60} & \best{62.00} \\
        Qwen2.5-3b 
          & 61.81 & 62.60 & 59.45 & $\equiv$
          & 77.56 & 78.39 & 78.17 & 76.35
          & 68.60 & 59.20 & 55.60 & 56.40\\
        Gemma2-2b 
          & 27.95 & 25.98 & \best{28.35} & 24.80
          & 61.64 & 62.62 & \best{62.85} & $\equiv$
          & 77.20 & 26.80 & 40.40 & 29.60 \\
        Granite3.2-2b 
          & 29.53 & 41.73 & \uline{42.91} & \best{43.31}
          & 59.89 & 63.84 & \uline{64.14} & \best{64.67}
          & 66.60 & 51.00 & 52.00 & 49.80 \\
        Granite3.3-2b 
          & 42.52 & 51.57 & \best{56.30} & \uline{55.51}
          & 64.37 & 69.52 & \best{73.62} & \uline{72.78}
          & 66.20 & 62.80 & 62.00 & 61.40 \\
        \bottomrule
      \end{tabular}
    }
\end{table*}

\begin{table*} [ht]
    \centering
    \caption{Results of all models on the StrategyQA, RACE-h, RACE-m, and TriviaQA datasets. The results are presented in zero-shot/zero-shot-CoT/Ours (static)/Ours (dynamic) format. \uline{Underlined} scores indicate when our approach outperforms the baselines but is not the best result.}
    \label{Oresults2}
    \resizebox{0.9\textwidth}{!}{
      \begin{tabular}{lcccccccccccccccc}
        \toprule
        & \multicolumn{4}{c}{\textbf{StrategyQA}} 
        & \multicolumn{4}{c}{\textbf{RACE-h}} 
        & \multicolumn{4}{c}{\textbf{RACE-m}} 
        & \multicolumn{4}{c}{\textbf{TriviaQA}} \\
        \cmidrule(lr{.5em}){2-5} 
        \cmidrule(lr{.5em}){6-9} 
        \cmidrule(lr{.5em}){10-13} 
        \cmidrule(lr{.5em}){14-17}
        \textbf{Model}
          & \textbf{0S} & \textbf{0CoT} & \textbf{Ours (S)} & \textbf{Ours (D)} 
          & \textbf{0S} & \textbf{0CoT} & \textbf{Ours (S)} & \textbf{Ours (D)} 
          & \textbf{0S} & \textbf{0CoT} & \textbf{Ours (S)} & \textbf{Ours (D)} 
          & \textbf{0S} & \textbf{0CoT} & \textbf{Ours (S)} & \textbf{Ours (D)} \\
        \midrule
        GPT-4.1 
          & 83.14 & 84.37 & \best{84.89} & \uline{84.54} 
          & 81.07 & 91.91 & \uline{91.97} & \best{92.08}
          & 89.28 & 94.99 & \uline{95.19} & \best{95.33}
          & 86.20 & 86.40 & \uline{87.40} & \best{87.50} \\
        GPT-4.1-mini 
          & 79.30 & 81.09 & \best{82.27} & \uline{81.66}
          & 84.19 & 89.37 & \best{90.37} & \uline{89.68}
          & 89.69 & 93.18 & \best{94.08} & \uline{93.87}
          & 79.90 & 82.80 & \best{84.90} & \uline{84.50} \\
        GPT-4.1-nano 
          & 70.09 & 68.86 & \best{76.03} & \uline{74.67} 
          & 79.22 & 80.76 & \best{83.16} & \uline{82.53}
          & 86.77 & 88.58 & \best{90.32} & \uline{90.04}
          & 73.90 & 79.40 & \uline{80.00} & \best{80.30} \\
        GPT-4o 
          & 80.57 & 81.09 & \best{82.45} & \uline{82.27} 
          & 92.31 & 89.25 & 89.11 & 89.57 
          & 94.71 & 93.38 & 94.29 & 94.22 
          & 85.60 & 87.00 & \best{87.10} & 87.00 \\
        GPT-4o-mini 
          & 76.90 & 79.04 & \uline{79.30} & \best{79.34}
          & 87.25 & 86.59 & \best{87.34} & 86.54
          & 92.27 & 92.76 & \best{93.11} & 92.62 
          & 77.70 & 79.50 & \best{80.60} & \uline{79.90} \\
        Phi4-14b 
          & 76.11 & 76.99 & \best{77.90} & \uline{77.34} 
          & 72.30 & 81.33 & \best{84.08} & \uline{83.73} 
          & 80.36 & 90.53 & \uline{90.60} & \best{90.67}
          & 73.10 & 75.90 & 74.60 & 75.70 \\
        Qwen2.5-14b 
          & 70.74 & 71.83 & \best{73.71} & \uline{73.54}
          & 85.25 & 83.79 & \best{87.65} & \uline{86.82}
          & 90.88 & 88.58 & \uline{91.43} & \best{91.78} 
          & 73.10 & 73.70 & \uline{75.20} & \best{75.50} \\
        Gemma3-12b 
          & 73.58 & 75.55 & 61.53 & \best{76.42} 
          & 23.70 & 57.55 & \uline{71.33} & \best{84.02}
          & 43.52 & 75.35 & \uline{88.37} & \best{88.79} 
          & 65.70 & 66.20 & \best{73.60} & \uline{71.30} \\
        Mistral-nemo-12b 
          & 61.62 & 68.34 & \uline{70.31} & \best{71.40} 
          & 52.86 & 53.46 & \uline{57.03} & \best{58.20}
          & 61.56 & 68.11 & \best{70.75} & \uline{69.99} 
          & 71.20 & 73.40 & \uline{74.50} & \best{74.60} \\
        Gemma2-9b 
          & 71.22 & 69.21 & 62.79 & 67.12 
          & 72.56 & 70.75 & \best{82.45} & \uline{81.42} 
          & 84.89 & 84.19 & \best{89.21} & \uline{88.68}
          & 69.70 & 69.30 & \best{70.70} & 66.50 \\
        Granite3.2-8b 
          & 60.35 & 63.49 & \best{65.28} & \uline{65.15} 
          & 70.90 & 67.10 & 68.47 & 65.07 
          & 77.65 & 77.23 & 76.74 & 73.47 
          & 68.60 & 71.30 & 70.60 & 69.80 \\
        Granite3.3-8b 
          & 63.01 & 66.16 & \uline{67.12} & \best{67.69} 
          & 70.07 & 70.78 & \uline{72.38} & \best{72.44} 
          & 76.67 & 81.89 & 79.81 & 80.50
          & 67.30 & 67.50 & 66.40 & 66.70 \\
        Llama3.1-8b 
          & 64.45 & 61.05 & \best{69.91} & \uline{69.69}
          & 75.33 & 65.81 & \uline{78.10} & \best{78.22} 
          & 82.10 & 76.39 & \best{84.68} & \uline{83.70}
          & 67.90 & 74.50 & \uline{74.60} & \best{75.40} \\
        Mistral-7b 
          & 39.56 & 38.78 & 36.11 & 38.47 
          & 30.42 & 26.76 & 24.73 & 29.00 
          & 40.53 & 42.06 & \best{43.18} &\uline{42.90} 
          & 53.20 & 54.30 & 46.90 & 46.60 \\
        Qwen2.5-7b 
          & 65.41 & 67.29 & \best{67.95} & 6.42
          & 82.39 & 79.93 & 81.45 & 82.02
          & 86.49 & 87.19 & \best{88.79} & \best{88.79} 
          & 59.90 & 64.20 & \uline{64.60} & \best{64.90} \\
        Llama3.2-3b 
          & 61.18 & 64.98 & \uline{66.90} & \best{67.34}
          & 68.67 & 65.58 & 66.35 & 67.67
          & 75.63 & 75.70 & \uline{76.25} & \best{77.16} 
          & 61.60 & 65.00 & \best{66.50} & \uline{65.20} \\
        Qwen2.5-3b 
          & 55.28 & 57.69 & \best{60.92} & \uline{60.74} 
          & 75.93 & 66.95 & 71.04 & 73.38
          & 82.03 & 75.42 & 78.69 & 81.13 
          & 50.30 & 52.70 & \best{55.50} & \uline{55.20} \\
        Gemma2-2b 
          & 64.10 & 58.38 & 60.52 & 58.58 
          & 49.91 & 54.63 & 52.12 & 52.57
          & 67.13 & 67.55 & 63.79 & 66.50 
          & 53.70 & 46.70 & 52.70 & 50.60 \\
        Granite3.2-2b 
          & 57.99 & 56.51 & 56.46 & 56.46
          & 65.87 & 58.35 & 57.72 & 58.43 
          & 71.52 & 65.81 & 63.51 & 63.51 
          & 54.30 & 55.30 & 54.70 & 54.40 \\
        Granite3.3-2b 
          & 61.31 & 59.56 & 58.08 & 59.17
          & 64.29 & 61.44 & 59.29 & 58.86 
          & 70.75 & 70.47 & 69.15 & 68.45 
          & 48.80 & 50.20 & \best{50.60} & 50.10 \\
        \bottomrule
      \end{tabular}
    }
  \end{table*}

Performance metrics of large LLMs across benchmarks highlight the differential efficacy of zero-shot, zero-shot-CoT, and our RE-ZS-CoT approaches. Tables \ref{Oresults1} and \ref{Oresults2} present the performance metrics of all 20 LLMs across seven benchmarks. To enhance readability, we showcase only the optimal role and prompt variant for each model-benchmark pair, facilitating a direct comparison with the baseline methodologies. The models are ordered starting with the GPT-4.1 and GPT-4o series, followed by open-weight models ranked by parameter size in descending order from 14b to 2b. Each result is presented in the format: zero-shot/zero-shot-CoT/RE-ZS-CoT(static)/RE-ZS-CoT(dynamic). Our RE-ZS-CoT framework demonstrates superior performance relative to both baselines across most benchmarks, with enhancements being particularly pronounced in arithmetic reasoning and reading comprehension datasets.\par

The RE-ZS-CoT framework consistently enhances arithmetic reasoning, with the most significant improvement observed in the open-weight model Gemma3-12b on the AQUA-RAT benchmark. Its accuracy surges from 23.2\% (zero-shot-CoT) to 67.3\% (static), a remarkable 44 percentage point (pp) increase, which is the most substantial gain recorded in this study. However, the same model's performance on the OpenBookQA benchmark decreases sharply, from a zero-shot baseline of 85.39\% to 51.40\% with our static approach. The closed-source GPT-4.1 model illustrates the paradigm’s potential, achieving a 7.2 pp gain on OpenBookQA (from 87.2\% to 94.4\%) while experiencing only a minor 1.2 pp decrease on GSM8K. This trend of mixed results is also visible in mid-sized open models; for instance, Qwen2.5-14b improves by nearly 2 pp on StrategyQA but declines by 5.6 pp on OpenBookQA. Conversely, the lightweight Mistral-7b improves its AQUA-RAT score by 4 pp and its GSM8K score by nearly 9 pp, yet its performance degrades on commonsense reasoning benchmarks. Overall, on the OpenBookQA dataset, both versions of our approach perform poorly. We suspect this is because of using zero-shot-CoT wording in our approach, which also inherits the flaw of zero-shot-CoT, as it did not perform consistently on common sense reasoning tasks, as shown in \cite{kojima2022large}.    \par

A strong correspondence emerges between the static and dynamic role selection versions; benchmarks where static roles yielded improvements were almost always those where dynamic roles also outperformed baselines. While static role selection often achieved peak performance, particularly on arithmetic tasks, the dynamic approach frequently proved competitive or superior on reading comprehension and commonsense reasoning tasks. For example, static roles consistently deliver the highest scores on AQUA-RAT for most models, whereas dynamic roles often achieve the best results on the RACE-h and StrategyQA benchmarks. This suggests that while pre-defined expert roles are highly effective for structured tasks, allowing models to autonomously select roles offers an adaptive advantage in more nuanced domains. When aggregating all model-task pairs, RE-ZS-CoT outperforms the stronger of the two baselines in approximately two-thirds of instances, with the most significant advantages found in arithmetic and reading comprehension datasets. \par

 \subsection{Nonconventional Roles}

\begin{table*}[t]
    \caption{Results for poet, comedian, and politician on AQuA-RAT, OpenBookQA, and TriviaQA benchmarks. Displayed as RE-ZS-CoT/auxiliary roles.}
    \label{NcNC}
    \centering
    \resizebox{0.8\textwidth}{!}{
    \begin{tabular}{lccc}
    \toprule
    & \multicolumn{3}{c}{\textbf{Poet}} \\
      \cmidrule(lr{.5em}){2-4} 
      \textbf{Model}  & \textbf{AQuA-RAT} & \textbf{OpenBookQA} & \textbf{TriviaQA}  \\
      \midrule
      GPT-4.1-nano & \best{77.95}/37.01 & \best{91.40}/87.06 & \best{80.30}/78.50  \\
      GPT-4o-mini & \best{78.74}/20.08 & \best{84.80}/75.00 & \best{80.60}/77.80     \\
      Gemma2-9b & \best{55.51}/3.94 & 36.80/\best{39.00}  & \best{70.70}/63.60 \\
      Granite3.2-8b & \best{60.63}/38.98 & \best{66.60}/53.00 &  \best{70.60}/65.60 \\
      Llama3.2-3b & \best{51.97}/35.04 & \best{62.00}/45.00 & \best{66.50}/61.00\\
      Qwen2.5-3b & \best{59.45}/22.05 & \best{56.40}/36.80 & \best{55.50}/42.70 \\
      \bottomrule
    \end{tabular}
    \begin{tabular}{ccc}
    \toprule
    \multicolumn{3}{c}{\textbf{Comedian}} \\
    \cmidrule(lr{.5em}){1-3} 
      \textbf{AQuA-RAT} &  \textbf{OpenBookQA} & \textbf{TriviaQA}  \\
      \midrule
      \best{77.95}/68.90 & \best{91.40}/81.60 &  \best{80.30}/75.70  \\
      \best{78.74}/76.77 & \best{84.80}/75.80 &  \best{80.60}/70.70     \\
      \best{55.51}/40.55 & \best{36.80}/5.20 &  \best{70.70}/57.20 \\
      \best{60.63}/42.91 & \best{66.60}/54.20 &  \best{70.60}/66.40 \\
      \best{51.97}/33.86 & \best{62.00}/41.80 &  \best{66.50}/59.20\\
      \best{59.45}/28.74 & \best{56.40}/21.00 &  \best{55.50}/44.70 \\
      \bottomrule
    \end{tabular} 
    \begin{tabular}{ccc}
    \toprule
     \multicolumn{3}{c}{\textbf{Politician}} \\
    \cmidrule(lr{.5em}){1-3} 
       \textbf{AQuA-RAT} & \textbf{OpenBookQA} & \textbf{TriviaQA}  \\
      \midrule
       \best{77.95}/78.74 & \best{91.40}/88.20  & \best{80.30}/77.50  \\
       78.74/\best{79.53} &  \best{84.80}/81.60 & \best{80.60}/78.50     \\
       \best{55.51}/31.50 & \best{36.80}/20.40 &  \best{70.70}/50.60 \\
       \best{60.63}/51.57 & \best{66.60}/48.20  & \best{70.60}/69.30 \\
       \best{51.97}/45.28 & \best{62.00}/54.20  & \best{66.50}/62.60\\
       \best{59.45}/47.24 & \best{56.40}/36.20 & \best{55.50}/54.20  \\
      \bottomrule
    \end{tabular}
    }
  \end{table*}
  
 Additionally, we conducted further experiments utilizing six LLMs (GPT-4.1-nano4, GPT-4o-mini, Gemma2:9b-instruct \cite{Gemma2}, Granite3.2:8b-instruct \cite{Granite32}, Qwen2.5:3b-instruct \cite{Qwen2.5}, and Llama3.2:3b-instruct \cite{meta2024Llama32}). These experiments were performed on three distinct datasets: AQuA-RAT, OpenbookQA, and TriviaQA. The selection of these six models aimed to illustrate the effects across varying model sizes, while the datasets were chosen to represent the impacts on all three categories. Namely, we tested three unconventional roles \textit{(Poet, Comedian, and Politician)} on these datasets. \par

We exclusively assessed the prompt variant and role, regardless of whether it was static or dynamic, that yielded superior performance for each LLM across the four selected datasets. Table \ref{NcNC} presents the outcomes of this experiment. We observed that unconventional roles diminish performance across all LLMs with only two exceptions. The decline in performance associated with unconventional roles was also noted in prior research \cite{kong2024better}.

\subsection{Role-Phrase Position and Static Self-Alignment Details}
\label{sec:B}
 \paragraph{Role-Phrase Position}

  \begin{figure*}[ht]
  \centering
    \includegraphics[width=.5\linewidth]{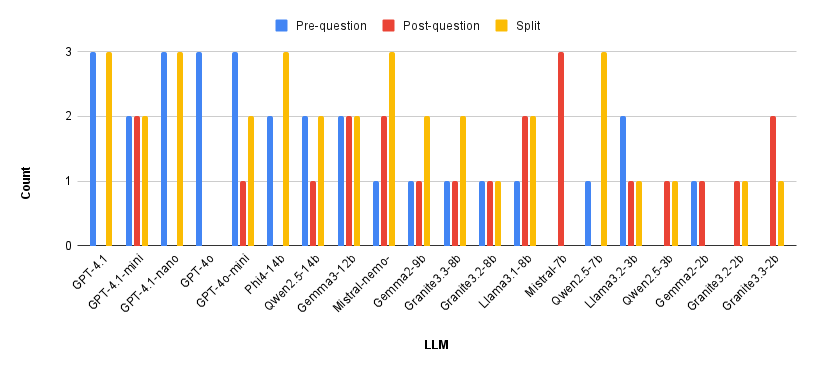}\hfill
    \includegraphics[width=.5\linewidth]{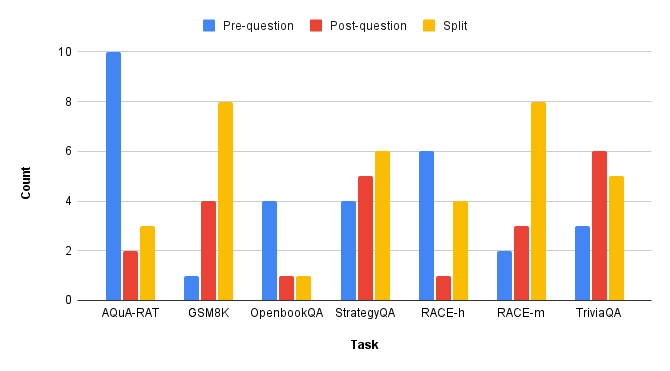}\hfill
    
  \caption{Role-phrase position results: comprehensive analysis of total counts across all benchmarks and all LLMs, respectively.}
  \label{fig:pp}
\end{figure*}
 We identify the most suitable prompt variants for different tasks by examining cases where they outperformed the baselines across benchmarks. The results presented in Figure \ref{fig:pp} indicate that the placement of the role-phrase position is contingent upon both the LLM and the specific benchmark, highlighting that a singular fixed prompt template may not be optimal. The data indicates that no single variant is consistently superior; performance is highly dependent on both the specific LLM and the benchmark task. For instance, the pre-question variant excels on AQuA-RAT, while the split variant leads on GSM8K. Given this variability, a fixed prompt placement is inadvisable. Instead, the role-phrase position should be treated as a key hyperparameter to be optimized for each specific model and task combination.

  \paragraph{ Effect of Roles and Self-Alignment}

   \begin{figure*}[ht]
  \centering
  \includegraphics[width=.50\linewidth]{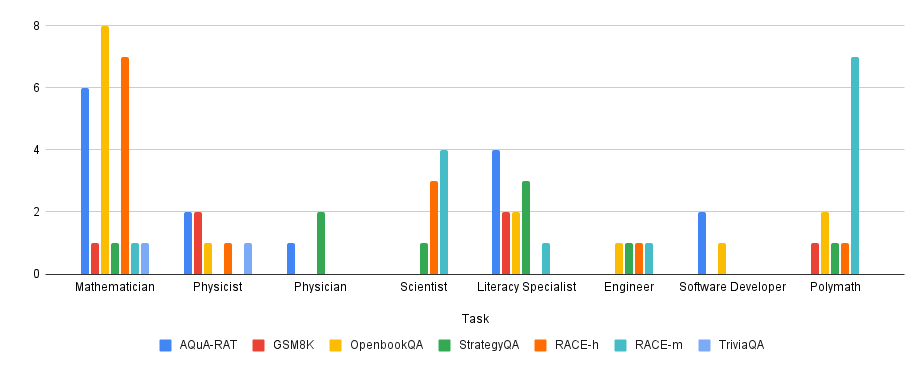}\hfill
  \includegraphics[width=.50\linewidth]{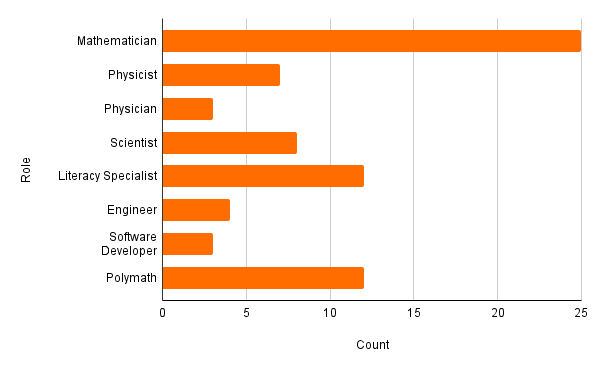}\hfill

  \caption{Effect of roles across all benchmarks.}
\label{fig:ER}
\end{figure*}

 In order to evaluate the most suitable role for various tasks, independent of the role-phrase position, we quantify the total occurrences in which each role exhibited superior performance compared to the baseline zero-shot and zero-shot-CoT across all benchmarks for all LLMs. Additionally, we demonstrate the impact of model self-alignment by presenting the instances in which the two selected roles surpassed the baselines. Figure \ref{fig:ER} illustrates this information, depicting the performance metrics of each role across all tasks, and the total role count for each role that exhibited superior performance, respectively. It is observed that roles deemed more appropriate for the benchmark, such as mathematicians for arithmetic reasoning, demonstrate superior performance. Overall, the roles of Mathematician, Literacy Specialist, and Polymath appear to yield the best performance.

\section{Ablations}
 \label{sec:6}
\paragraph{Role-enhanced Zero-shot}

\begin{table*}[t]
    \caption{Results for our Role-enhanced zero-shot on AQuA-RAT, OpenBookQA, and TriviaQA benchmarks. Displayed as zero-shot/RE-ZS.}
    \label{RE-ZS}
    \centering
    \resizebox{.5\textwidth}{!}{
  \begin{tabular}{lccc}
    \toprule
      \textbf{Model}  & \textbf{AQuA-RAT} & \textbf{OpenBookQA.} & \textbf{TriviaQA}  \\
      \midrule
      GPT-4.1-nano & 75.98/\best{79.53}  & \best{87.40}/84.40 & 73.90/\best{76.00}  \\
      GPT-4o-mini & 67.32/\best{79.92} & \best{92.60}/87.20 & 77.70/\best{78.50} \\
      Gemma2-9b & 54.72/54.72  & 88.60/\best{88.80} & 69.70/\best{70.70} \\
      Granite3.2-8b & 53.54/\best{56.69}  & \best{66.80}/65.20 & 68.60/\best{70.50} \\
      Llama3.2-3b & \best{55.91}/54.72 & 30.20/\best{49.80} & 61.60/\best{66.10} \\
      Qwen2.5-3b & \best{61.81}/60.63 & \best{68.60}/54.80 & 50.30/\best{52.10} \\
      \bottomrule
    \end{tabular}
    }
  \end{table*}

We evaluated a zero-shot variant of our role-enhanced prompt (RE-ZS), utilizing the aforementioned LLMs and datasets during the additional experimentation segment of Section 4, following the procedure categorized under the Nonconventional Roles. An example of this zero-shot variant is illustrated in Fig.\ref{examplep}(c). The results of this experiment are presented in Table \ref{RE-ZS}. We observed an improvement in performance with our zero-shot approach, where the models performed best on the TriviaQA dataset. In addition, we noted an enhancement in performance compared to RE-ZS-CoT in the OpenBookQA dataset using Gemma2-9b and Qwen2.5-3b. Furthermore, our zero-shot strategy outperformed both zero-shot and zero-shot-CoT on multiple occasions, highlighting the effectiveness of role prompting.

\paragraph{Lexical Variations}

\begin{table*}[t]
    \caption{Results for lexical variations on AQuA-RAT, OpenBookQA, and TriviaQA benchmarks. Displayed as the relative performance increase or decrease from the RE-ZS-CoT.}
    \label{LeVa}
    \centering
    \resizebox{0.9\textwidth}{!}{
    \begin{tabular}{lccccccccc}
      \toprule
          & \multicolumn{3}{c}{\textbf{AQuA-RAT}} 
          & \multicolumn{3}{c}{\textbf{OpenBookQA}} 
         & \multicolumn{3}{c}{\textbf{TriviaQA}} \\
          \cmidrule(lr{.5em}){2-4} 
          \cmidrule(lr{.5em}){5-7}
        \cmidrule(lr{.5em}){8-10} 
      \textbf{Model} 
            & \textbf{w/o ``Expert"} & \textbf{``Top"} & \textbf{``Amateur"} 
            & \textbf{w/o ``Expert"} & \textbf{``Top"} & \textbf{``Amateur"}  
            & \textbf{w/o ``Expert"} & \textbf{``Top"} & \textbf{``Amateur"} \\
      \midrule
      GPT-4.1-nano & 0 \% &  \best{0.40\%}  & \textcolor{red}{-0.39\%} &
      \textcolor{red}{-1.84\%} & \textcolor{red}{-5.34\%} & \textcolor{red}{-6.24\%}
      & \textcolor{red}{-7.24\%} & \textcolor{red}{-0.10\%} & \textcolor{red}{-1.00\%} \\
      GPT-4o-mini & \textcolor{red}{-1.18\%} & \textcolor{red}{-0.51\%} & \textcolor{red}{-5.79\%}
      & \textcolor{red}{-2.03\%} & \textcolor{red}{-2.33\%} & \textcolor{red}{-3.93\%}
      &  \textcolor{red}{-1.10\%} & \textcolor{red}{-0.80\%} & \textcolor{red}{-2.90\%} \\
      Gemma2-9b
          & \textcolor{red}{-4.57\%}
          & \textcolor{red}{-4.18\%}
          & \textcolor{red}{-4.97\%}
          & \textcolor{red}{-0.56\%}
          & \textcolor{red}{-1.25\%}
          & \textcolor{red}{-1.44\%}
          &  \textcolor{red}{-1.40\%}
          & \textcolor{red}{-1.80\%}
          & \textcolor{red}{-3.80\%}\\
  
      Granite3.2-8b
          & \textcolor{red}{-3.15\%}
          & \textcolor{red}{-3.94\%}
          & \textcolor{red}{-4.15\%}
          & \textcolor{red}{-4.4\%}
          & \textcolor{red}{-2.4\%}
          & \textcolor{red}{-5.7\%}
          &  0 \%
          & \textcolor{red}{-0.90\%}
          & \textcolor{red}{-1.30\%}\\
  
      Llama3.2-3b
          & \textcolor{red}{-1.97\%}
          & \best{3.54\%}
          & \textcolor{red}{-2.39\%}
          & \textcolor{red}{-12.00\%}
          & \textcolor{red}{-18.2\%}
          &  \textcolor{red}{-20.6\%}
          &  \textcolor{red}{-1.80\%}
          & \textcolor{red}{-2.30\%}
          & \textcolor{red}{-3.20\%} \\
  
      Qwen2.5-3b
          & \best{3.15\%}
          & \textcolor{red}{-3.94\%}
          & \textcolor{red}{-4.2\%}
          &  \textcolor{red}{-5.90\%}
          & \textcolor{red}{-2.3\%}
          & \textcolor{red}{-8.60\%} 
          &  \textcolor{red}{-0.50\%}
          & \textcolor{red}{-0.70\%}
          & \textcolor{red}{-1.00\%} \\
  
      \bottomrule
    \end{tabular}
    }
  \end{table*}

In addition to unconventional roles, role combination, and role-enhanced zero-shot, we also evaluated our approach utilizing three distinct lexical variations. These variations encompass the removal of the term "expert” from the prompt, substituting "expert" with the terms "top" and "amateur." This methodology is employed to assess the robustness of the performance effects that roles exert on a task and to validate the utilization of the term "expert" in our prompt. We employ the term "amateur" to investigate whether performance is negatively influenced, as previous research indicates \cite{kong2024better}. Consistent with the other experiments, we utilized the same six models and four datasets. Table \ref{LeVa} illustrates the results of the lexical variations. The performance of our RE-ZS-CoT is best when used with the term "expert"; a decrease in performance is evident when it is removed or altered to terms other than a few exceptions, where improvement is observed when the term "expert" is changed. These improvements are limited to the AQuA-RAT dataset, where "top" demonstrated enhancement in two instances. The performance of all LLMs diminishes when utilizing the term "amateur" more so than other variations, thereby reinforcing the findings of previous research \cite{kong2024better}.

\section{Discussion}
 \label{sec:7}
Our experiments show that role-enhanced prompting boosts zero-shot reasoning across tasks and model sizes. RE-ZS-CoT biases LLMs towards domain-specific reasoning. This aligns with previous research on plan-and-solve strategies \cite{zhou2023plan}. The placement of the role phrase (pre-question, post-question, split-phrase) affects performance, and variability across models and benchmarks necessitates adaptive prompt engineering rather than rigid templates.  Unconventional roles led to lower performance, revealing that LLMs struggle with conflicting or irrelevant expertise. Lexical variation experiments confirm that terms like "expert" in role prompts are significant; substituting them with "amateur" or omitting them lowered accuracy, reinforcing that explicit expertise cues guide LLMs toward structured, factual reasoning.

Our experiments reveal an interesting interaction between the role self-alignment strategy and the nature of the reasoning task. Static role self-alignment demonstrates superior efficacy in highly structured domains that benefit from specialized knowledge. This is most pronounced on the AQUA-RAT benchmark, where leveraging a specific role like "Mathematician" consistently yields peak performance. Conversely, the dynamic role selection mechanism, which permits the model to infer the most suitable persona autonomously, often proves more advantageous for tasks requiring broader or more adaptive reasoning. On commonsense reasoning benchmarks, such as StrategyQA, and reading comprehension tasks, like RACE (h\&m), the dynamic approach frequently leads to improved outcomes, suggesting that flexibility in role adoption is beneficial when the reasoning path is less prescribed. These findings suggest a fundamental trade-off: static roles maximize performance when a task's requirements align perfectly with a known expert archetype, but this specialization can be suboptimal for more multifaceted problems. Dynamic selection, while not always reaching the same performance ceiling on specialized tasks, offers a more robust and generalizable strategy, mitigating the risk of a misaligned persona.

The rapid emergence of test-time compute and natively reasoning models raises a critical question: Does role-based zero-shot prompting still hold value when models are trained to reason automatically via reinforcement learning? Our findings indicate that it does, primarily as a lightweight, resource-efficient alternative. Native reasoning architectures utilize extensive test-time compute, making them costly, token-heavy, and slow, which limits their viability for latency-sensitive or large-scale deployments. RE-ZS-CoT offers a practical solution by biasing standard LLMs towards domain-specific reasoning without the massive inference-time costs associated with RL-driven search mechanisms. Notably, the success of RE-ZS-CoT on smaller, open-weight models ranging from 2b to 14b parameters demonstrates that adaptive role-imposition can simulate specialized analytical thinking. This effectively democratizes access to high-quality reasoning in resource-constrained settings  where deploying a heavy-compute System 2 model is unfeasible.

\paragraph{Ethical considerations}
The proposed RE-ZS-CoT possesses the potential to democratize access to high-quality reasoning in LLMs. By enhancing zero-shot performance, our methodology diminishes the reliance on curated examples, thereby increasing the accessibility of LLMs in low-resource environments. However, augmenting these capabilities may also exacerbate risks, including the generation of persuasive misinformation by priming the model with a role that excels at this task. While this study primarily concentrates on technical advancements, it is imperative to consider safeguards such as output validation and ethical guidelines to mitigate potential misuse. Furthermore, domain-specific roles may inherit biases from the training data, necessitating the establishment of rigorous evaluation frameworks to ensure fairness and transparency.

\paragraph{Limitations}
While RE-ZS-CoT demonstrates broad empirical success, our study has several important limitations. First, the lack of transparency in LLM training constrains our work. Many models, especially closed-source ones like GPT-4.1 and GPT-4o, lack public documentation on their training datasets. This opacity makes reproducibility difficult and obscures whether the observed improvements stem from genuine role-based reasoning pathways or latent memorization. Consequently, this limits our ability to conduct comprehensive fairness and bias evaluations, as the composition of the training data heavily influences how a model defines and operationalizes a given "expert" role.

Second, while our massive-scale evaluation highlights the variance in RE-ZS-CoT's efficacy, most notably its degraded performance on generalized commonsense tasks like OpenBookQA, we do not conduct a deep qualitative or mechanistic error analysis. Our current study successfully identifies when and where the method struggles, but leaves the precise how and why unanswered. We hypothesize that rigid role constraints may induce overthinking in intuitive contexts, but further interpretability research is required to map exactly how persona adoption shifts internal representations of reasoning.

Third, our study relies on the broad empirical consensus generated by testing 20 distinct LLMs across seven datasets, reporting absolute accuracy gains rather than conducting paired statistical significance testing (e.g., McNemar's test) for every individual model-task pair. While macro-trends and double-digit percentage surges (e.g., on AQuA-RAT) provide robust evidence of structural shifts in reasoning, smaller single-digit improvements on datasets with limited sample sizes should be interpreted as competitive performance rather than definitive algorithmic superiority.

Fourth, our evaluations exclusively focus on standard autoregressive language models. With the recent paradigm shift toward test-time compute and natively reasoning models (System 2 thinkers) that utilize reinforcement learning to automatically generate search and reasoning traces (e.g., OpenAI o-series, DeepSeek-R1), the interaction between explicit zero-shot role prompting and implicit RL-driven reasoning remains unexplored. Finally, our experimental scope is constrained to English text-only corpora. Evaluating the efficacy and cultural nuances of RE-ZS-CoT in non-English corpora, conversational multi-turn tasks, and multimodal environments remains an important area for future exploration

Future studies need to focus on RE-ZS-CoT through several perspectives. Firstly, testing our approach with the two-stage prompting method along with RS-few-shot will reveal the extent of its usefulness. Secondly, examining multi-role collaboration will evaluate whether using specialized personas can create a division of cognitive labor that offers benefits beyond single-role configurations. Finally, we will test RE-ZS-CoT in areas like code generation, multimodal understanding, conversational tasks, and non-English corpora to determine its applicability beyond current standards.
\section{Conclusion}

We presented RE-ZS-CoT, a role-enhanced prompting framework that significantly improved zero-shot reasoning in LLMs. Our method achieved enhanced performance across seven benchmarks and 20 models by systematically integrating domain-specific roles with CoT prompting, with accuracy gains of up to 44 percentage points. The findings highlighted the critical role of domain alignment in prompt design. Moreover, the efficiency of the framework and scalability made it ideal for low-resource settings. By advancing zero-shot reasoning without reliance on fine-tuning or multi-turn interactions, RE-ZS-CoT bridged the gap between generic prompting and domain-specific expertise. This study highlights the transformative potential of role-aware prompting while also highlighting the importance of role-phrase position.

\bibliography{custom}

@inproceedings{wei2022chain,
	title        = {{Chain-of-Thought Prompting Elicits Reasoning in Large Language Models}},
	author       = {Wei, Jason and Wang, Xuezhi and Schuurmans, Dale and Bosma, Maarten and Ichter, Brian and Xia, Fei and Chi, Ed and Le, Quoc and Zhou, Denny},
	year         = 2022,
	booktitle    = {Advances in Neural Information Processing Systems},
	volume       = 35,
	pages        = {24824--24837}
}

@inproceedings{kojima2022large,
	title        = {{Large Language Models are Zero-Shot Reasoners}},
	author       = {Kojima, Takeshi and Gu, Shixiang Shane and Reid, Machel and Matsuo, Yutaka and Iwasawa, Yusuke},
	year         = 2022,
	booktitle    = {Advances in Neural Information Processing Systems},
	volume       = 35,
	pages        = {22166--22177}
}

@inproceedings{wang2024chain,
	title        = {{Chain-of-Thought Reasoning Without Prompting}},
	author       = {Wang, Xuezhi and Zhou, Denny},
	year         = 2024,
	booktitle    = {Advances in Neural Information Processing Systems},
	volume       = 38
}

@inproceedings{zhou2023plan,
	title        = {{Plan-and-Solve Prompting: Improving Zero-Shot Chain-of-Thought Reasoning by Large Language Models}},
	author       = {Zhou, Yujie and Wang, Yizhong and Zhang, Shuyan and Liu, Ximing and Chen, Xiang and Wang, Yuwei and Zhou, Denny},
	year         = 2023,
	booktitle    = {Proceedings of the 61st Annual Meeting of the Association for Computational Linguistics (ACL)},
	pages        = {1879--1892}
}

@inproceedings{besta2024graph,
	title        = {{Graph of Thoughts: Solving Elaborate Problems with Large Language Models}},
	author       = {Besta, Maciej and Blach, Nils and Kubicek, Ales and others},
	year         = 2024,
	booktitle    = {Proceedings of the AAAI Conference on Artificial Intelligence},
	volume       = 38,
	number       = 16,
	pages        = {17711--17719}
}

@inproceedings{salewski2023context,
	title        = {{In-Context Impersonation Reveals Large Language Models’ Strengths and Biases}},
	author       = {Salewski, Lukas and Alaniz, Samuel and Rio-Torto, Inês and Schulz, Eric and Akata, Zeynep},
	year         = 2023,
	booktitle    = {Advances in Neural Information Processing Systems},
	volume       = 36
}

@inproceedings{kong2024better,
	title        = {{Better Zero-Shot Reasoning with Role-Play Prompting}},
	author       = {Kong, Aoran and Zhao, Shijie and Chen, Hao and Li, Qichen and Qin, Yong and Sun, Rui and Zhou, Xin and Wang, En and Dong, Xin},
	year         = 2024,
	booktitle    = {Proceedings of the 2024 Conference of the North American Chapter of the Association for Computational Linguistics (NAACL)},
	pages        = {3456--3468}
}

@inproceedings{yuan2024instance,
	title        = {{Instance-adaptive Zero-shot Chain-of-Thought Prompting}},
	author       = {Yuan, Xiaosong and others},
	year         = 2024,
	booktitle    = {Advances in Neural Information Processing Systems},
	volume       = 38
}

@inproceedings{sun2023principle,
	title        = {{Principle-Driven Self-Alignment of Language Models from Scratch}},
	author       = {Sun, Zhiqing and Shen, Yikang and Zhou, Qinhong and Zhang, Hongxin and Chen, Zhenfang and Cox, David and Yang, Yiming and Gan, Chuang},
	year         = 2023,
	booktitle    = {Advances in Neural Information Processing Systems},
	volume       = 36
}

@article{wang2022self,
	title        = {{Self-Instruct: Aligning Language Models with Self-Generated Instructions}},
	author       = {Wang, Yizhong and Kordi, Yeganeh and Mishra, Swaroop and Liu, Alisa and Smith, Noah A and Khashabi, Daniel and Hajishirzi, Hannaneh},
	year         = 2023,
	journal      = {Proceedings of the 61st Annual Meeting of the Association for Computational Linguistics (Volume 1: Long Papers)}
}

@inproceedings{madaan2023self,
	title        = {{Self-Refine: Iterative Refinement with Self-Feedback}},
	author       = {Madaan, Aman and others},
	year         = 2023,
	booktitle    = {Advances in Neural Information Processing Systems},
	volume       = 36
}

@article{sanh2021multitask,
	title        = {{Multitask Prompted Training Enables Zero-Shot Task Generalization}},
	author       = {Sanh, Victor and Webson, Albert and Raffel, Colin and Bach, Stephen H and Sutawika, Lintang and Alyafeai, Zaid and Chaffin, Antoine and Stiegler, Arnaud and Scao, Teven Le and Raja, Arun and others},
	journal      = {Tenth International Conference on Learning Representations, {ICLR}},
        year         = 2022
}

@article{wang2022self2,
	title        = {{Self-Consistency Improves Chain of Thought Reasoning in Language Models}},
	author       = {Wang, Xuezhi and Wei, Jason and Schuurmans, Dale and Le, Quoc and Chi, Ed and Narang, Sharan and Chowdhery, Aakanksha and Zhou, Denny},
	journal      = {The Eleventh International Conference on Learning Representations,{ICLR} 2023, Kigali, Rwanda, May 1-5, 2023}
}

@inproceedings{reynolds2021prompt,
	title        = {{Prompt Programming for Large Language Models: Beyond the Few-Shot Paradigm}},
	author       = {Reynolds, Laria and McDonell, Kyle},
	year         = 2021,
	booktitle    = {{Extended Abstracts of the 2021 CHI Conference on Human Factors in Computing Systems}},
	pages        = {1--7}
}

@article{kim2023cot,
        title        = {{The {C}o{T} Collection: Improving Zero-shot and Few-shot Learning of Language Models via Chain-of-Thought Fine-Tuning}},
        author       = {Kim, Seungone  and Joo, Se  and Kim, Doyoung  and Jang, Joel  and Ye, Seonghyeon  and Shin, Jamin  and Seo, Minjoon},
        journal      = {{Proceedings of the 2023 Conference on Empirical Methods in Natural Language Processing}},
        year         = 2023,
        pages        = "12685--12708",
}

@inproceedings{cao2024worst,
	title        = {{On the Worst Prompt Performance of Large Language Models}},
	author       = {Cao, Bowen and others},
	year         = 2024,
	booktitle    = {Advances in Neural Information Processing Systems},
	volume       = 38
}

@article{shanahan2023role,
        title        = {{Role Play with Large Language Models}},
        author       = {Shanahan, Murray and McDonell, Kyle and Reynolds, Laria},
        journal      = {Nature},
        year         = {2023}
}

@inproceedings{zheng2024when,
	title        = {{When {\textquotedblright}A Helpful Assistant{\textquotedblright} Is Not Really Helpful: Personas in System Prompts Do Not Improve Performances of Large Language Models}},
	author       = {Zheng, Mingqian and Pei, Jiaxin and Logeswaran, Lajanugen and Lee, Moontae and Jurgens, David},
	year         = 2024,
	booktitle    = {Findings of the Association for Computational Linguistics: EMNLP 2024},
	pages        = {15126--15154}
}

@inproceedings{Aqua,
	title        = {{Program Induction by Rationale Generation: Learning to Solve and Explain Algebraic Word Problems}},
	author       = {Ling, Wang  and Yogatama, Dani  and Dyer, Chris  and Blunsom, Phil},
	year         = 2017,
	month        = jul,
	booktitle    = {Proceedings of the 55th Annual Meeting of the Association for Computational Linguistics (Volume 1: Long Papers)},
	publisher    = {Association for Computational Linguistics},
	address      = {Vancouver, Canada},
	pages        = {158--167}
}

@article{gsm8k,
	title        = {{Training Verifiers to Solve Math Word Problems}},
	author       = {Cobbe, Karl and Kosaraju, Vineet and Bavarian, Mohammad and Chen, Mark and Jun, Heewoo and Kaiser, Lukasz and Plappert, Matthias and Tworek, Jerry and Hilton, Jacob and Nakano, Reiichiro and Hesse, Christopher and Schulman, John},
	year         = 2021,
	journal      = {arXiv preprint arXiv:2110.14168}
}

@article{strategyqa,
	title        = {{Did Aristotle Use a Laptop? A Question Answering Benchmark with Implicit Reasoning Strategies}},
	author       = {Geva, Mor and Khashabi, Daniel and Segal, Elad and Khot, Tushar and Roth, Dan and Berant, Jonathan},
	year         = 2021,
	month        = {04},
	journal      = {Transactions of the Association for Computational Linguistics},
	volume       = 9,
	pages        = {346--361}
}

@inproceedings{OpenBookQA2018,
	title        = {{Can a Suit of Armor Conduct Electricity? A New Dataset for Open Book Question Answering}},
	author       = {Todor Mihaylov and Peter Clark and Tushar Khot and Ashish Sabharwal},
	year         = 2018,
	booktitle    = {Empirical Methods in Natural Language Processing, {EMNLP}}
}

@inproceedings{race,
	title        = {{{RACE}: Large-scale {R}e{A}ding Comprehension Dataset From Examinations}},
	author       = {Lai, Guokun  and Xie, Qizhe  and Liu, Hanxiao  and Yang, Yiming  and Hovy, Eduard},
	year         = 2017,
	month        = sep,
	booktitle    = {Proceedings of the 2017 Conference on Empirical Methods in Natural Language Processing, {EMNLP}},
	address      = {Copenhagen, Denmark},
	pages        = {785--794}
}

@inproceedings{triviaqa,
	title        = {{{T}rivia{QA}: A Large Scale Distantly Supervised Challenge Dataset for Reading Comprehension}},
	author       = {Joshi, Mandar  and Choi, Eunsol  and Weld, Daniel  and Zettlemoyer, Luke},
	year         = 2017,
	month        = jul,
	booktitle    = {Proceedings of the 55th Annual Meeting of the Association for Computational Linguistics (Volume 1: Long Papers)},
	publisher    = {Association for Computational Linguistics},
	address      = {Vancouver, Canada},
	pages        = {1601--1611}
}

@article{phi4,
	title        = {{Phi-4 Technical Report}},
	author       = {Abdin, Marah and Aneja, Jyoti and Behl, Harkirat and Bubeck, S{\'e}bastien and Eldan, Ronen and Gunasekar, Suriya and Harrison, Michael and Hewett, Russell J and Javaheripi, Mojan and Kauffmann, Piero and others},
	year         = 2024,
	journal      = {arXiv preprint arXiv:2412.08905}
}

@misc{mistralnemo,
	title        = {{Mistral-Nemo-Instruct-2407}},
	author       = {Mistral AI and NVIDIA},
	year         = 2024
}

@article{qwen2.5,
	title        = {{Qwen2.5 Technical Report}},
	author       = {Yang, An and Yang, Baosong and Zhang, Beichen and Hui, Binyuan and Zheng, Bo and Yu, Bowen and Li, Chengyuan and Liu, Dayiheng and Huang, Fei and Wei, Haoran and others},
	year         = 2024,
	journal      = {arXiv preprint arXiv:2412.15115}
}

@article{gemma3,
	title        = {{Gemma 3 Technical Report}},
	author       = {Team, Gemma and Kamath, Aishwarya and Ferret, Johan and Pathak, Shreya and Vieillard, Nino and Merhej, Ramona and Perrin, Sarah and Matejovicova, Tatiana and Ram{\'e}, Alexandre and Rivi{\`e}re, Morgane and others},
	year         = 2025,
	journal      = {arXiv preprint arXiv:2503.19786}
}

@misc{granite32,
	title        = {{IBM Granite 3.2: Open-Source Reasoning and Vision Models}},
	author       = {{IBM Research}},
	year         = 2025
}

@misc{granite33,
	title        = {{IBM Granite 3.3: Enhanced Reasoning and Multimodal Capabilities}},
	author       = {{IBM Research}},
	year         = 2025
}

@misc{meta2024llama32,
	title        = {{Llama 3.2: Multimodal and Lightweight Large Language Models}},
	author       = {{Meta AI}},
	year         = 2024
}

@misc{jiang2023mistral7b,
	title        = {{Mistral 7B}},
	author       = {Albert Q. Jiang and Alexandre Sablayrolles and Arthur Mensch and Chris Bamford and Devendra Singh Chaplot and others},
	year         = 2023,
	eprint       = {2310.06825},
	archiveprefix = {arXiv},
	primaryclass = {cs.CL}
}

@article{llama3.1,
	title        = {{The Llama 3 Herd of Models}},
	author       = {Grattafiori, Aaron and Dubey, Abhimanyu and Jauhri, Abhinav and Pandey, Abhinav and Kadian, Abhishek and Al-Dahle, Ahmad and Letman, Aiesha and Mathur, Akhil and Schelten, Alan and Vaughan, Alex and others},
	year         = 2024,
	journal      = {arXiv preprint arXiv:2407.21783}
}

@article{gemma2,
	title        = {{Gemma 2: Improving Open Language Models at a Practical Size}},
	author       = {Team, Gemma and Riviere, Morgane and Pathak, Shreya and Sessa, Pier Giuseppe and Hardin, Cassidy and Bhupatiraju, Surya and Hussenot, L{\'e}onard and Mesnard, Thomas and Shahriari, Bobak and Ram{\'e}, Alexandre and others},
	year         = 2024,
	journal      = {arXiv preprint arXiv:2408.00118}
}

@misc{ollama2025,
  author       = {{Ollama Inc.}},
  title        = {{Ollama}},
  year         = {2025},
  note         = {Accessed: 2025-05-15}
}

@article{zheng2023helpful,
  title={Is" a helpful assistant" the best role for large language models? a systematic evaluation of social roles in system prompts},
  author={Zheng, Mingqian and Pei, Jiaxin and Jurgens, David},
  journal={arXiv preprint arXiv:2311.10054},
  volume={8},
  year={2023}
}

@inproceedings{wang2024role,
  title={Role prompting guided domain adaptation with general capability preserve for large language models},
  author={Wang, Rui and Mi, Fei and Chen, Yi and Xue, Boyang and Wang, Hongru and Zhu, Qi and Wong, Kam-Fai and Xu, Ruifeng},
  booktitle={Findings of the Association for Computational Linguistics: NAACL 2024},
  pages={2243--2255},
  year={2024}
}

@article{wang2025improving,
  title={Improving llm reasoning through interpretable role-playing steering},
  author={Wang, Anyi and Shu, Dong and Wang, Yifan and Ma, Yunpu and Du, Mengnan},
  journal={arXiv preprint arXiv:2506.07335},
  year={2025}
}

@inproceedings{han2024rethinking,
  title={Rethinking the role-play prompting in mathematical reasoning tasks},
  author={Han, Zhiguang and Wang, Zijian},
  booktitle={Proceedings of the 1st Workshop on Efficiency, Security, and Generalization of Multimedia Foundation Models},
  pages={13--17},
  year={2024}
}

@article{jaech2024openai,
  title={Openai o1 system card},
  author={Jaech, Aaron and Kalai, Adam and Lerer, Adam and Richardson, Adam and El-Kishky, Ahmed and Low, Aiden and Helyar, Alec and Madry, Aleksander and Beutel, Alex and Carney, Alex and others},
  journal={arXiv preprint arXiv:2412.16720},
  year={2024}
}

@article{guo2025deepseek,
  title={Deepseek-r1: Incentivizing reasoning capability in llms via reinforcement learning},
  author={Guo, Daya and Yang, Dejian and Zhang, Haowei and Song, Junxiao and Wang, Peiyi and Zhu, Qihao and Xu, Runxin and Zhang, Ruoyu and Ma, Shirong and Bi, Xiao and others},
  journal={arXiv preprint arXiv:2501.12948},
  year={2025}
}
\bibliographystyle{IEEEtran}

\appendix

\section{Additional Experiment Details}
\label{sec:A}
\subsection{Code}
Code, along with results, is available at: \url{https://github.com/ZoltharZay/RE-ZS-CoT} 

\subsection{Datasets}
We utilized publicly accessible datasets, the specifics of which are provided in Table \ref{tab:dataset-details}. Below, we provide a description of each dataset.

\begin{table}[ht]
\centering
\caption{Details of each dataset. \textbf{Test split} shows the split used for evaluations in this research, while \textbf{Count} shows the number of included samples. For the \textbf{Task}, AR: Arithmetic Reasoning, CR: Commonsense Reasoning, RC: Reading Comprehension.\\
\textsuperscript{*}TriviaQA was downsampled to only 1000 records to save computational resources.}
\label{tab:dataset-details}
\begin{tabular}{llll}
\toprule
\textbf{Dataset} & \textbf{Task} & \textbf{Test split} & \textbf{Count}  \\
\midrule
AQuA-RAT   & AR   & test        & 254      \\
GSM8K      & AR   & test        & 1319   \\
OepnbookQA & CR   & dev         & 500     \\
StrategyQA & CR   & test        & 2290     \\
RACE-m     & RC   & middle-test & 1436     \\
RACE-h     & RC   & high-test   & 3498     \\
TriviaQA   & RC  & Wikipedia   & 1000\textsuperscript{*}  \\
\bottomrule
\end{tabular}
\end{table}

\paragraph{AQuA-RAT}
AQUA-RAT (Algebra Question Answering with Rationales) is a comprehensive dataset containing approximately 100,000 algebraic word problems. Each question's solution is meticulously documented in a step-by-step format utilizing natural language. This dataset is employed to train a program generation model that acquires the ability to produce both the explanatory narrative and the program designed to resolve the respective question.
\paragraph{GSM8K}
GSM8K (Grade School Math 8K) constitutes a dataset comprising 8,500 high-quality, linguistically diverse grade school mathematics word problems. This dataset has been developed to facilitate the task of answering questions related to fundamental mathematical issues that necessitate multi-step reasoning. The problems within this dataset require between two and eight steps to resolve. Solutions predominantly entail executing a series of elementary calculations employing basic arithmetic operations (addition, subtraction, multiplication, and division) to derive the final answer.
\paragraph{OpenBookQA}
OpenBookQA seeks to advance research in sophisticated question-answering methodologies, facilitating a deeper comprehension of both the subject matter (with salient facts encapsulated as an open book, accompanied by the dataset) and the linguistic constructs utilized to articulate it. Specifically, it encompasses inquiries that necessitate multi-step reasoning, the application of supplementary common and commonsense knowledge, as well as comprehensive text interpretation. OpenBookQA represents a novel category of question-answering dataset, designed in accordance with the principles of open-book examinations for evaluating human proficiency in a given subject.
 \paragraph{StrategyQA}
 \textit{StrategyQA} constitutes a benchmark for question answering, wherein the requisite reasoning steps are inherently embedded within the question, necessitating inference through a strategic approach. This benchmark comprises a total of 2,780 examples, each encompassing a strategic question, its corresponding decomposition, and supporting evidence paragraphs. The questions featured in StrategyQA are concise, encompass a diverse array of topics, and address a broad spectrum of strategies.
 \paragraph{RACE(h\&m)}
  \textit{RACE} represents a comprehensive dataset for reading comprehension comprising over 28,000 passages and nearly 100,000 associated questions. This dataset has been curated from English examinations administered in China, specifically targeting middle school and high school students.
\paragraph{TriviaQA}
\textit{TriviaQA} is a reading comprehension dataset that comprises over 650,000 question-answer-evidence triples. The dataset includes 95,000 question-answer pairs created by trivia enthusiasts, along with independently sourced evidence documents, an average of six per question, that provide high-quality distant supervision for answering the questions.

The subsequent list delineates the sources of data utilized for this study.
\begin{enumerate}
    \item AQuA-RAT: \url{https://github.com/google-deepmind/AQuA}
    \item GSM8K: \url{https://github.com/openai/grade-school-math}
    \item OpenbookQA: \url{https://github.com/allenai/OpenBookQA}
    \item StrategyQA: \url{https://github.com/google/BIG-bench/tree/main/bigbench/benchmark_tasks/strategyqa}
    \item RACE(h\&m): \url{https://www.cs.cmu.edu/~glai1/data/race/}
    \item TriviaQA: \url{https://nlp.cs.washington.edu/triviaqa/}

\end{enumerate}

\subsection{Models and Implementation Details}
\label{sec:A.2}
We conduct experiments utilizing a total of 20 LLMs ranging from 2b to 14b open-weights models along with large closed-source models from OpenAI. This exploration aims to investigate the impact of our approach on varying parameter sizes, extending from small to large. For the main experiments, we use all 20 models, which are listed as follows: GPT-4.1-2025-04-14, GPT-4.1-mini-2025-04-14, GPT-4.1-nano-2025-04-14, GPT-4o-2024-08-06, GPT-4o-mini-2024-07-18, Phi4:14b-q4\_K\_M \cite{Phi4}, Qwen2.5:14b-instruct-q4\_K\_M \cite{Qwen2.5}, Gemma3:12b-it-q4\_K\_M \cite{Gemma3}, Mistral-nemo:12b-instruct-2407-q4\_K\_M \cite{Mistralnemo}, Gemma2:9b-instruct-q4\_K\_M \cite{Gemma2}, Granite3.2:8b-instruct-q4\_K\_M \cite{Granite32}, Granite3.3:8b \cite{Granite33}, Llama3.1:8b-text-q4\_K\_M \cite{Llama3.1}, Mistral:7b-instruct-q4\_K\_M \cite{jiang2023Mistral7b}, Qwen2.5:7b-instruct-q4\_K\_M \cite{Qwen2.5}, Llama3.2:3b-instruct-q4\_K\_M  \cite{meta2024Llama32}, Qwen2.5:3b-instruct-q4\_K\_M \cite{Qwen2.5}, Granite3.2:2b-instruct-q4\_K\_M \cite{Granite32}, Granite3.3:2b \cite{Granite33}, and Gemma2:2b-text-q4\_K\_M \cite{Gemma2}.\par

We explicitly list the complete names of all models, encompassing the closed-source OpenAI models, to ensure traceability in the event of future updates. For models with open weights, we utilized the open-source platform Ollama \cite{ollama2025}. Ollama facilitates users in executing LLMs directly on their local machines. It provides both a command-line interface and a REST API, and also integrates with the OpenAI API to enhance functionality. We employed the OpenAI API for both closed-source GPT models and models with open weights, while setting the sampling temperature to zero and providing a fixed random seed. Consequently, we interact with the model solely once for each record of the dataset, following a prompting approach. Any supplementary settings, including safety measures, are maintained at their default values. \par

Subsequent to the aforementioned details, all open-weight models were obtained for local execution via Ollama. Our experiments were conducted on two machines: one equipped with an NVIDIA A100 80GB and the other with two NVIDIA RTX 6000 Ada 48GB GPUs. The total number of GPU hours required to complete all experiments is approximately 5600 (NVIDIA A100). The budget allocated for querying the OpenAI models amounts to approximately \$2200.


\section{Additional Arithmetic Reasoning Dataset}
To further verify whether the static role self-alignment works effectively on arithmetic reasoning, we tested our approach against two baselines on the MATH dataset. The Mathematics Aptitude Test of Heuristics dataset, abbreviated as MATH, comprises 12,500 mathematical problems. The MATH dataset encompasses problems derived from various mathematics competitions, including the AMC 10, AMC 12, AIME, among others. These competitions have been conducted over several decades and evaluate the mathematical problem-solving capabilities of the most distinguished young talent in the field of mathematics. We used the test set consisting of 5000 questions. \par
\begin{enumerate}
 \item MATH: \url{https://github.com/hendrycks/math/?tab=MIT-1-ov-file}
 \end{enumerate}

The MATH dataset is further divided into seven categories, as it encompasses a wide range of topics. These categories are Pre-Algebra, Algebra, Number Theory, Counting and Probability, Intermediate Algebra, Geometry, and Pre-Calculus. For this dataset, we also added instructions for the model to output in a specified format to aid in the evaluation process, since it is a more complex dataset than the others.

 Table \labelcref{OresultsM1} illustrates the performance across the seven distinct topics previously mentioned. To enhance readability and clarity, we showcase only the optimal role and variant for each LLM on a benchmark, facilitating comparisons with zero-shot and zero-shot-CoT methodologies just like in the main manuscript. The results show that our RE-ZS-CoT consistently outperforms the two baselines. These results solidify the finding that for a structured task like arithmetic reasoning, a specific role, such as "Mathematician", works best, as the top-selected role by all LLMs out of the eight roles for this task was "Mathematician".
\begin{table*}[ht]
    \centering
    \caption{Results of all models on the MATH benchmark. The results are presented in zero-shot/zero-shot-CoT/RE-ZS-CoT(static) format.}
    \label{OresultsM1}
    \resizebox{\textwidth}{!}{
      \begin{tabular}{lcccccccccccccccccccccccccccccc}
        \toprule
        & \multicolumn{3}{c}{\textbf{Algebra}} 
        & \multicolumn{3}{c}{\textbf{Count. \& Prob}} 
        & \multicolumn{3}{c}{\textbf{Geometry}} 
        & \multicolumn{3}{c}{\textbf{Int. Algebra}} 
        & \multicolumn{3}{c}{\textbf{Num. Th.}}
        & \multicolumn{3}{c}{\textbf{Prealgebra}} 
        & \multicolumn{3}{c}{\textbf{Precalculus}} \\
        \cmidrule(lr{.5em}){2-4} 
        \cmidrule(lr{.5em}){5-7} 
        \cmidrule(lr{.5em}){8-10} 
        \cmidrule(lr{.5em}){11-13} 
        \cmidrule(lr{.5em}){14-16} 
        \cmidrule(lr{.5em}){17-19} 
        \cmidrule(lr{.5em}){20-22}
        \textbf{Model}
          & \textbf{0S} & \textbf{0CoT} & \textbf{Ours(S)} 
          & \textbf{0S} & \textbf{0CoT} & \textbf{Ours(S)} 
          & \textbf{0S} & \textbf{0CoT} & \textbf{Ours(S)} 
          & \textbf{0S} & \textbf{0CoT} & \textbf{Ours(S)} 
          & \textbf{0S} & \textbf{0CoT} & \textbf{Ours(S)} 
          & \textbf{0S} & \textbf{0CoT} & \textbf{Ours(S)} 
          & \textbf{0S} & \textbf{0CoT} & \textbf{Ours(S)} \\
        \midrule
        GPT-4.1
          & 88.12 & 87.87 & \best{88.54}
          & 79.75 & 81.01 & 80.59
          & 70.98 & 69.73 & \best{73.31}
          & 71.87 & 72.20 & \best{73.09}
          & 91.30 & 90.19 & 91.11
          & 83.01 & 83.81 & 83.70
          & 63.74 & 65.75 & 64.10 \\
        GPT-4.1-mini
          & 90.06 & 90.56 & \best{91.07}
          & 81.43 & 81.01 & \best{83.33}
          & 72.86 & 74.74 & \best{75.78}
          & 80.07 & 78.96 & \best{81.95}
          & 91.85 & 92.22 & 92.04
          & 85.53 & 84.50 & 85.42
          & 74.91 & 72.71 & \best{74.91} \\
        GPT-4.1-nano
          & 85.93 & 86.86 & \best{87.45}
          & 70.68 & 70.46 & \best{72.78}
          & 61.17 & 60.54 & \best{62.63}
          & 65.01 & 66.67 & 66.11
          & 85.19 & 84.07 & \best{87.04}
          & 78.30 & 78.87 & \best{81.29}
          & 56.78 & 62.09 & 61.90 \\
        GPT-4o
          & 87.53 & 86.86 & \best{88.15}
          & 69.83 & 70.89 & \best{72.36}
          & 58.66 & 57.83 & \best{60.33}
          & 57.59 & 55.91 & 57.25
          & 79.44 & 79.07 & \best{81.67}
          & 80.02 & 81.06 & \best{82.93}
          & 54.03 & 52.56 & \best{57.46} \\
        GPT-4o-mini
          & 84.92 & 84.92 & \best{86.10}
          & 67.09 & 64.56 & 66.67
          & 54.70 & 56.99 & 56.16
          & 51.94 & 52.38 & \best{54.15}
          & 78.89 & 78.52 & \best{79.63}
          & 78.42 & 80.25 & \best{80.94}
          & 49.08 & 49.63 & 49.08 \\
        Phi4-14b
          & 87.87 & 88.12 & \best{88.29}
          & 68.14 & 67.93 & \best{71.73}
          & 62.42 & 62.00 & 62.00
          & 57.81 & 59.80 & \best{60.80}
          & 80.56 & 79.63 & \best{81.67}
          & 80.48 & 79.68 & \best{80.83}
          & 57.51 & 56.59 & \best{57.69} \\
        Qwen2.5-14b
          & 85.85 & 84.67 & \best{85.85}
          & 62.87 & 64.35 & \best{65.19}
          & 56.16 & 54.49 & 54.70
          & 54.93 & 53.71 & \best{55.15}
          & 72.59 & 72.59 & \best{74.26}
          & 77.61 & 76.81 & \best{77.96}
          & 55.13 & 53.66 & 54.58 \\
        Gemma3-12b
          & 86.10 & 85.68 & \best{86.60}
          & 66.03 & 65.19 & \best{66.88}
          & 52.40 & 55.74 & \best{56.37}
          & 16.05 & 15.17 & \best{27.02}
          & 65.74 & 65.56 & 58.33
          & 73.59 & 72.45 & 67.97
          & 19.60 & 20.33 & 18.50 \\
        Mistral-nemo-12b
          & 49.12 & 50.63 & \best{53.41}
          & 32.91 & 30.80 & 31.43
          & 23.80 & 28.81 & \best{29.44}
          & 14.17 & 16.72 & \best{17.83}
          & 26.30 & 29.44 & \best{30.19}
          & 50.63 & 54.42 & \best{57.52}
          & 22.53 & 21.25 & 21.79 \\
        Gemma2-9b
          & 64.20 & 63.94 & \best{65.21}
          & 32.07 & 33.54 & 32.91
          & 31.73 & 32.57 & \best{34.03}
          & 17.72 & 17.50 & \best{18.94}
          & 38.52 & 36.48 & \best{40.74}
          & 60.39 & 61.31 & \best{61.42}
          & 27.66 & 26.37 & \best{27.84} \\
        Granite3.2-8b
          & 62.93 & 59.39 & 62.01
          & 38.19 & 34.39 & 35.65
          & 34.03 & 31.11 & \best{35.07}
          & 24.25 & 23.48 & 23.92
          & 39.81 & 35.93 & 38.70
          & 59.36 & 56.49 & 57.75
          & 28.94 & 28.39 & \best{30.59} \\
        Granite3.3-8b
          & 71.86 & 77.59 & 76.24
          & 43.88 & 45.78 & \best{49.79}
          & 40.71 & 43.84 & \best{46.56}
          & 39.65 & 38.21 & \best{40.64}
          & 55.74 & 59.63 & \best{62.41}
          & 62.69 & 66.59 & \best{68.54}
          & 41.03 & 43.96 & 42.31 \\
        Llama3.1-8b
          & 62.51 & 64.53 & \best{64.62}
          & 37.97 & 38.19 & \best{42.83}
          & 35.91 & 32.99 & 35.49
          & 24.03 & 22.04 & 22.26
          & 39.63 & 41.30 & \best{42.22}
          & 61.54 & 62.00 & \best{62.00}
          & 24.36 & 24.54 & \best{27.84} \\
        Mistral-7b
          & 8.17 & 8.68 & \best{9.60}
          & 6.75 & 8.86 & \best{8.86}
          & 2.51 & 3.13 & \best{3.66}
          & 1.88 & 1.11 & \best{2.44}
          & 5.93 & 3.70 & \best{6.48}
          & 11.48 & 12.86 & \best{16.07}
          & 1.83 & 2.56 & \best{2.75} \\
        Qwen2.5-7b
          & 84.50 & 82.06 & 83.24
          & 60.13 & 58.02 & \best{60.97}
          & 54.49 & 50.52 & 52.61
          & 50.61 & 47.51 & 49.72
          & 68.33 & 65.74 & \best{69.26}
          & 75.20 & 74.17 & \best{76.46}
          & 49.82 & 49.63 & 48.72 \\
        Llama3.2-3b
          & 39.09 & 57.29 & \best{58.13}
          & 23.00 & 32.07 & 30.59
          & 19.21 & 28.18 & \best{29.23}
          & 14.62 & 19.16 & \best{20.82}
          & 27.04 & 35.93 & \best{36.30}
          & 40.07 & 54.54 & 52.35
          & 15.75 & 25.27 & 24.54 \\
        Qwen2.5-3b
          & 73.80 & 72.96 & 72.03
          & 43.67 & 42.83 & 42.19
          & 43.84 & 40.71 & 41.54
          & 36.77 & 36.99 & 34.55
          & 48.52 & 52.41 & 50.00
          & 64.06 & 66.02 & 65.56
          & 37.36 & 36.63 & 36.81 \\
        Gemma2-2b
          & 31.09 & 29.23 & \best{32.27}
          & 11.60 & 7.17 & 9.70
          & 9.39 & 7.31 & 8.77
          & 5.32 & 4.54 & 5.20
          & 13.15 & 12.59 & \best{14.63}
          & 28.13 & 26.87 & \best{31.69}
          & 7.33 & 8.97 & \best{9.89} \\
        Granite3.2-2b
          & 26.20 & 39.51 & \best{40.78}
          & 15.40 & 19.62 & \best{23.21}
          & 14.55 & 21.50 & 21.29
          & 11.18 & 11.30 & \best{12.74}
          & 15.83 & 19.91 & \best{21.30}
          & 26.75 & 39.49 & 38.46
          & 15.93 & 15.75 & \best{16.85} \\
        Granite3.3-2b
          & 59.81 & 61.42 & \best{66.81}
          & 32.07 & 32.07 & \best{40.93}
          & 33.19 & 31.11 & \best{35.28}
          & 20.27 & 25.36 & \best{29.13}
          & 41.67 & 37.04 & \best{43.70}
          & 50.98 & 55.80 & \best{60.28}
          & 26.19 & 26.37 & \best{29.12} \\
        \bottomrule
      \end{tabular}
    }
  \end{table*}
 
\section{Cross-Validation of Role Self-Alignment}
\label{sec:D}

\subsection{Static Self-Alignment}
\begin{figure*}[ht]
  \centering
  \includegraphics[width=0.5\linewidth]{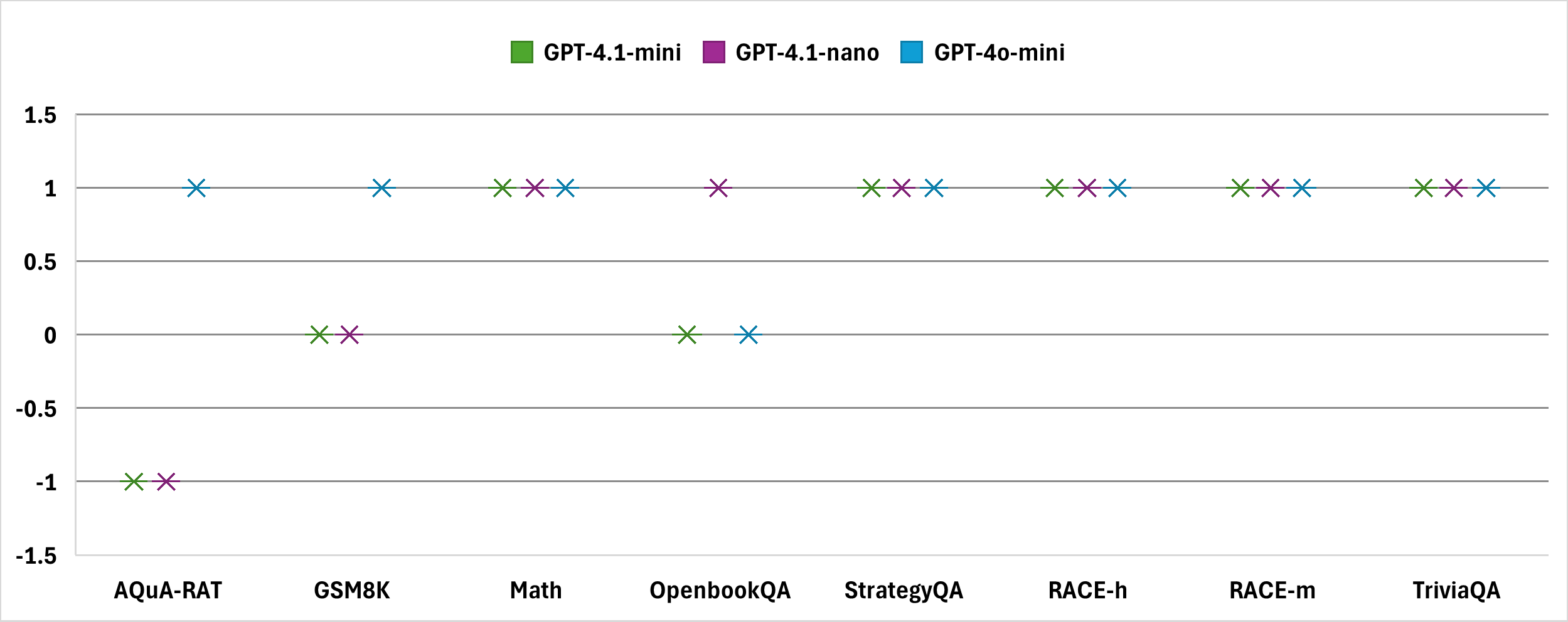}\hfill
  \includegraphics[width=0.5\linewidth]{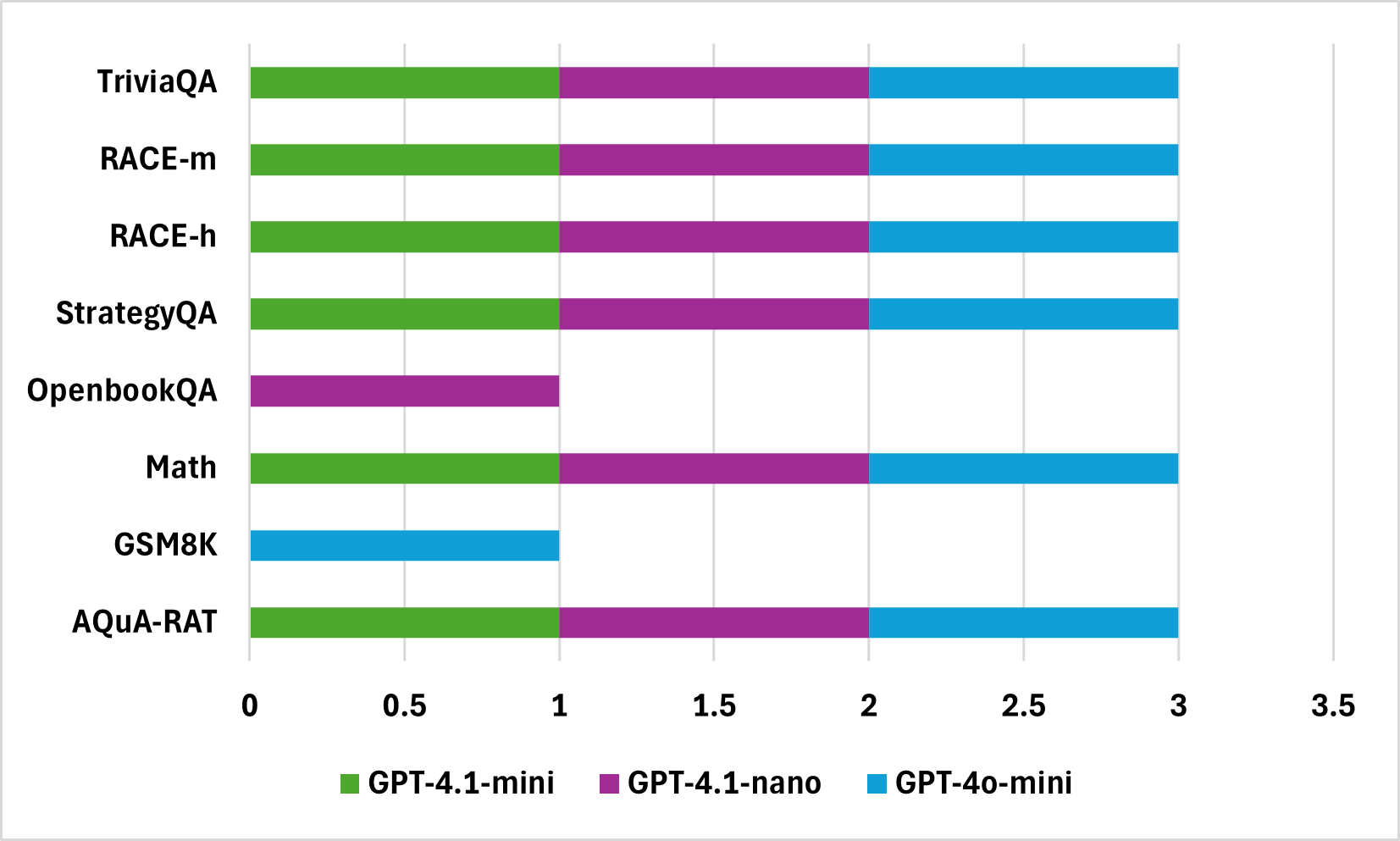}\hfill
  \caption{Effect of roles and self-alignment across all benchmarks.}
\label{fig:CVSA1}
\end{figure*}

In order to cross-validate the overall effectiveness of our static self-alignment approach, we conducted tests on the smaller and less expensive variants of the GPT models (GPT-4.1-mini, GPT-4.1-nano, and GPT-4o-mini). Subsequently, we directed these LLMs to select the most appropriate roles from a predetermined list based on the provided task description, performing this selection 100 times, analogous to the open-weight models. The objective was to assess whether the two best-performing roles for a given task were chosen by the model itself and to determine the frequency of these selections. Tables \labelcref{slgpt1,slgpt2,slgpt4,slgpt7,slgpt8,slgpt9} present the results of static self-alignment for the GPT models. Additionally, Figure \ref{fig:CVSA1} illustrates the instances in which the models outperformed the baselines, as well as depicts the correlation between the roles selected by the model and its performance on those roles, respectively. A value of 1 indicates that the model's two selected roles were the ones that surpassed the baseline; a value of 0 indicates no performance improvement compared to the baselines, while a value of -1 signifies a performance increase over the baseline, although the role is not one of the two selected roles. \par

The selected GPT models exhibit superior self-alignment in comparison to the open-weight models, indicating that larger models demonstrate enhanced self-alignment capabilities relative to smaller models. This highlights the necessity for improved self-alignment in smaller models. 

\begin{table*}[ht]
\centering
\caption{Static Role Self-Alignment on AQuA (GPT Models)}
\label{slgpt1}
\resizebox{1\textwidth}{!}{
\begin{tabular}{lrrrrrrrr}
\hline
Model & Mathematician & Physicist & Physician & Scientist & Literacy Specialist & Engineer & Software Developer & Polymath \\
\hline
GPT-4.1-mini & 100 & 9 & 0 & 0 & 1 & 2 & 88 & 0 \\ \hline
GPT-4.1-nano & 99 & 0 & 0 & 60 & 0 & 17 & 10 & 14 \\ \hline
GPT-4o-mini & 100 & 0 & 0 & 0 & 100 & 0 & 0 & 0 \\ \hline
\end{tabular} }
\end{table*}

\begin{table*}[ht]
\centering
\caption{Static Role Self-Alignment on GSM8K (GPT Models)}
\label{slgpt2}
\resizebox{1\textwidth}{!}{
\begin{tabular}{lrrrrrrrr}
\hline
Model & Mathematician & Physicist & Physician & Scientist & Literacy Specialist & Engineer & Software Developer & Polymath \\
\hline
GPT-4.1-mini & 100 & 5 & 0 & 13 & 0 & 39 & 30 & 13 \\ \hline
GPT-4.1-nano & 100 & 0 & 1 & 82 & 0 & 16 & 1 & 0 \\ \hline
GPT-4o-mini & 100 & 0 & 0 & 0 & 63 & 37 & 0 & 0 \\ \hline
\end{tabular} }
\end{table*}

\begin{table*}[ht]
\centering
\caption{Static Role Self-Alignment on OpenbookQA (GPT Models)}
\label{slgpt4}
\resizebox{1\textwidth}{!}{
\begin{tabular}{lrrrrrrrr}
\hline
Model & Mathematician & Physicist & Physician & Scientist & Literacy Specialist & Engineer & Software Developer & Polymath \\
\hline
GPT-4.1-mini & 7 & 8 & 0 & 92 & 4 & 0 & 0 & 89 \\ \hline
GPT-4.1-nano & 1 & 1 & 1 & 100 & 6 & 9 & 7 & 75 \\ \hline
GPT-4o-mini & 0 & 0 & 0 & 47 & 83 & 0 & 0 & 70 \\ \hline
\end{tabular} }
\end{table*}

\begin{table*}[ht]
\centering
\caption{Static Role Self-Alignment on StrategyQA (GPT Models)}
\label{slgpt7}
\resizebox{1\textwidth}{!}{
\begin{tabular}{lrrrrrrrr}
\hline
Model & Mathematician & Physicist & Physician & Scientist & Literacy Specialist & Engineer & Software Developer & Polymath \\
\hline
GPT-4.1-mini & 3 & 2 & 0 & 97 & 0 & 0 & 0 & 98 \\ \hline
GPT-4.1-nano & 0 & 0 & 0 & 100 & 0 & 10 & 1 & 89 \\ \hline
GPT-4o-mini & 2 & 0 & 0 & 25 & 70 & 4 & 0 & 99 \\ \hline
\end{tabular} }
\end{table*}

\begin{table*}[ht]
\centering
\caption{Static Role Self-Alignment on RACE(M\&h) (GPT Models)}
\label{slgpt8}
\resizebox{1\textwidth}{!}{
\begin{tabular}{lrrrrrrrr}
\hline
Model & Mathematician & Physicist & Physician & Scientist & Literacy Specialist & Engineer & Software Developer & Polymath \\
\hline
GPT-4.1-mini & 0 & 0 & 0 & 0 & 100 & 0 & 85 & 15 \\ \hline
GPT-4.1-nano & 0 & 0 & 0 & 33 & 100 & 0 & 17 & 49 \\ \hline
GPT-4o-mini & 0 & 0 & 0 & 4 & 100 & 0 & 90 & 6 \\ \hline
\end{tabular} }
\end{table*}

\begin{table*}[ht]
\centering
\caption{Static Role Self-Alignment on TriviaQA (GPT Models)}
\label{slgpt9}
\resizebox{1\textwidth}{!}{
\begin{tabular}{lrrrrrrrr}
\hline
Model & Mathematician & Physicist & Physician & Scientist & Literacy Specialist & Engineer & Software Developer & Polymath \\
\hline
GPT-4.1-mini & 5 & 5 & 0 & 16 & 76 & 0 & 36 & 62 \\ \hline
GPT-4.1-nano & 0 & 0 & 0 & 96 & 13 & 0 & 87 & 4 \\ \hline
GPT-4o-mini & 0 & 0 & 0 & 47 & 92 & 2 & 11 & 48 \\ \hline
\end{tabular} }
\end{table*}

We asked the open-weights LLMs themselves to select the most appropriate roles from a predetermined list based on the provided task description. Each LLM was queried one hundred times, and we selected the two roles most frequently identified for use in the experiments. Tables \labelcref{sl1,sl2,sl4,sl7,sl8,sl9} presents the comprehensive results of the self-alignment.

\begin{table*}[ht]
\centering
\caption{Static Role Self-Alignment on AQuA}
\label{sl1}
\resizebox{1\textwidth}{!}{
\begin{tabular}{lrrrrrrrr}
\hline
Model & Mathematician & Physicist & Physician & Scientist & Literacy Specialist & Engineer & Software Developer & Polymath \\
\hline
Phi4-14b & 100 & 12 & 0 & 0 & 17 & 0 & 66 & 5 \\ \hline
Qwen2.5-14b & 100 & 39 & 0 & 0 & 14 & 29 & 15 & 3 \\ \hline
Gemma3-12b & 4 & 0 & 0 & 0 & 0 & 0 & 99 & 97 \\ \hline
Mistral-nemo-12b & 95 & 0 & 0 & 7 & 82 & 2 & 13 & 1 \\ \hline
Gemma2-9b & 100 & 0 & 0 & 2 & 0 & 1 & 95 & 2 \\ \hline
Granite3.2-8b & 99 & 12 & 1 & 28 & 24 & 1 & 35 & 0 \\ \hline
Granite3.3-8b & 100 & 0 & 3 & 2 & 3 & 1 & 91 & 0 \\ \hline
Llama3.1-8b & 90 & 26 & 1 & 5 & 24 & 4 & 13 & 37 \\ \hline
Mistral-7b & 65 & 68 & 4 & 18 & 4 & 26 & 0 & 1 \\ \hline
Qwen2.5-7b & 98 & 19 & 2 & 9 & 40 & 2 & 28 & 2 \\ \hline
Llama3.2-3b & 45 & 13 & 4 & 28 & 71 & 6 & 1 & 32 \\ \hline
Qwen2.5-3b & 54 & 18 & 1 & 11 & 67 & 8 & 39 & 2 \\ \hline
Gemma2-2b & 98 & 5 & 0 & 7 & 84 & 0 & 4 & 2 \\ \hline
Granite3.2-2b & 92 & 41 & 4 & 21 & 18 & 10 & 10 & 4 \\ \hline
Granite3.3-2b & 99 & 3 & 7 & 2 & 83 & 2 & 1 & 0 \\ \hline
\end{tabular} }
\end{table*}

\begin{table*}[ht]
\centering
\caption{Static Role Self-Alignment on GSM8K}
\label{sl2}
\resizebox{1\textwidth}{!}{
\begin{tabular}{lrrrrrrrr}
\hline
Model & Mathematician & Physicist & Physician & Scientist & Literacy Specialist & Engineer & Software Developer & Polymath \\
\hline
Phi4-14b         & 100 & 9  & 0 & 4  & 11 & 9  & 38 & 29 \\
\hline
Qwen2.5-14b      & 100 & 15 & 0 & 0  & 0  & 78 & 5  & 2  \\
\hline
Gemma3-12b       & 97  & 0  & 0 & 0  & 0  & 0  & 17 & 86 \\
\hline
Mistral-nemo-12b & 94  & 1  & 0 & 1  & 32 & 34 & 32 & 6  \\
\hline
Gemma2-9b        & 100 & 0  & 0 & 7  & 76 & 1  & 14 & 2  \\
\hline
Granite3.2-8b    & 98  & 9  & 1 & 9  & 12 & 12 & 59 & 0  \\
\hline
Granite3.3-8b    & 100 & 1  & 0 & 2  & 12 & 10 & 75 & 0  \\
\hline
Llama3.1-8b      & 97  & 37 & 0 & 0  & 13 & 1  & 0  & 52 \\
\hline
Mistral-7b       & 77  & 57 & 2 & 19 & 2  & 28 & 3  & 1  \\
\hline
Qwen2.5-7b       & 100 & 23 & 0 & 28 & 1  & 24 & 7  & 17 \\
\hline
Llama3.2-3b      & 41  & 8  & 2 & 14 & 72 & 4  & 0  & 58 \\
\hline
Qwen2.5-3b       & 90  & 10 & 0 & 5  & 36 & 25 & 5  & 29 \\
\hline
Gemma2-2b        & 86  & 0  & 0 & 0  & 98 & 0  & 14 & 1  \\
\hline
Granite3.2-2b    & 91  & 46 & 1 & 20 & 14 & 18 & 4  & 6  \\
\hline
Granite3.3-2b    & 98  & 24 & 5 & 3  & 60 & 7  & 3  & 0  \\
\hline
\end{tabular} }
\end{table*}

\begin{table*}[ht]
\centering
\caption{Static Role Self-Alignment on OpenbookQA}
\label{sl4}
\resizebox{1\textwidth}{!}{
\begin{tabular}{lrrrrrrrr}
\hline
Model & Mathematician & Physicist & Physician & Scientist & Literacy Specialist & Engineer & Software Developer & Polymath \\
\hline
Phi4-14b & 27 & 20 & 1 & 22 & 51 & 0 & 2 & 77 \\ \hline
Qwen2.5-14b & 37 & 37 & 0 & 53 & 5 & 5 & 0 & 63 \\ \hline
Gemma3-12b & 0 & 0 & 0 & 100 & 0 & 0 & 0 & 100 \\ \hline
Mistral-nemo-12b & 0 & 0 & 1 & 88 & 25 & 3 & 0 & 83 \\ \hline
Gemma2-9b & 1 & 0 & 0 & 98 & 0 & 0 & 10 & 91 \\ \hline
Granite3.2-8b & 3 & 67 & 4 & 96 & 10 & 0 & 7 & 13 \\ \hline
Granite3.3-8b & 44 & 34 & 13 & 63 & 0 & 6 & 11 & 29 \\ \hline
Llama3.1-8b & 7 & 74 & 0 & 33 & 4 & 0 & 0 & 82 \\ \hline
Mistral-7b & 43 & 68 & 8 & 28 & 7 & 23 & 5 & 4 \\ \hline
Qwen2.5-7b & 11 & 20 & 2 & 86 & 38 & 1 & 0 & 42 \\ \hline
Llama3.2-3b & 7 & 52 & 12 & 63 & 13 & 6 & 1 & 44 \\ \hline
Qwen2.5-3b & 62 & 70 & 5 & 21 & 11 & 6 & 1 & 24 \\ \hline
Gemma2-2b & 29 & 10 & 2 & 86 & 15 & 4 & 1 & 50 \\ \hline
Granite3.2-2b & 79 & 48 & 5 & 18 & 8 & 12 & 3 & 27 \\ \hline
Granite3.3-2b & 10 & 11 & 2 & 25 & 9 & 39 & 12 & 90 \\ \hline
\end{tabular} }
\end{table*}

\begin{table*}[ht]
\centering
\caption{Static Role Self-Alignment on StrategyQA}
\label{sl7}
\resizebox{1\textwidth}{!}{
\begin{tabular}{lrrrrrrrr}
\hline
Model & Mathematician & Physicist & Physician & Scientist & Literacy Specialist & Engineer & Software Developer & Polymath \\
\hline
Phi4-14b & 51 & 12 & 1 & 56 & 0 & 4 & 17 & 59 \\ \hline
Qwen2.5-14b & 34 & 16 & 2 & 3 & 0 & 79 & 40 & 26 \\ \hline
Gemma3-12b & 0 & 0 & 0 & 94 & 0 & 0 & 6 & 100 \\ \hline
Mistral-nemo-12b & 0 & 5 & 0 & 82 & 3 & 9 & 14 & 87 \\ \hline
Gemma2-9b & 24 & 0 & 0 & 99 & 0 & 32 & 35 & 10 \\ \hline
Granite3.2-8b & 1 & 35 & 3 & 91 & 12 & 1 & 33 & 24 \\ \hline
Granite3.3-8b & 83 & 6 & 3 & 37 & 0 & 20 & 14 & 36 \\ \hline
Llama3.1-8b & 24 & 62 & 2 & 58 & 2 & 8 & 2 & 42 \\ \hline
Mistral-7b & 49 & 66 & 6 & 24 & 1 & 28 & 4 & 2 \\ \hline
Qwen2.5-7b & 14 & 10 & 1 & 75 & 6 & 5 & 1 & 88 \\ \hline
Llama3.2-3b & 21 & 49 & 1 & 80 & 1 & 9 & 0 & 39 \\ \hline
Qwen2.5-3b & 65 & 49 & 6 & 31 & 8 & 8 & 8 & 25 \\ \hline
Gemma2-2b & 78 & 35 & 7 & 44 & 0 & 17 & 8 & 11 \\ \hline
Granite3.2-2b & 65 & 54 & 4 & 20 & 7 & 23 & 6 & 19 \\ \hline
Granite3.3-2b & 45 & 19 & 7 & 23 & 12 & 22 & 11 & 60 \\ \hline
\end{tabular} }
\end{table*}

\begin{table*}[ht]
\centering
\caption{Static Role Self-Alignment on RACE(M\&h)}
\label{sl8}
\resizebox{1\textwidth}{!}{
\begin{tabular}{lrrrrrrrr}
\hline
Model & Mathematician & Physicist & Physician & Scientist & Literacy Specialist & Engineer & Software Developer & Polymath \\
\hline
Phi4-14b & 13 & 9 & 2 & 38 & 69 & 2 & 63 & 2 \\ \hline
Qwen2.5-14b & 2 & 2 & 0 & 0 & 98 & 0 & 95 & 3 \\ \hline
Gemma3-12b & 5 & 8 & 1 & 70 & 95 & 13 & 7 & 1 \\ \hline
Mistral-nemo-12b & 16 & 5 & 1 & 30 & 97 & 2 & 4 & 45 \\ \hline
Gemma2-9b & 9 & 0 & 0 & 46 & 100 & 1 & 44 & 0 \\ \hline
Granite3.2-8b & 65 & 60 & 13 & 25 & 11 & 10 & 3 & 10 \\ \hline
Granite3.3-8b & 10 & 1 & 13 & 18 & 100 & 10 & 46 & 0 \\ \hline
Llama3.1-8b & 3 & 13 & 0 & 84 & 84 & 3 & 12 & 1 \\ \hline
Mistral-7b & 2 & 1 & 0 & 61 & 100 & 17 & 11 & 8 \\ \hline
Qwen2.5-7b & 36 & 27 & 4 & 36 & 65 & 10 & 18 & 1 \\ \hline
Llama3.2-3b & 2 & 6 & 1 & 9 & 100 & 0 & 23 & 59 \\ \hline
Qwen2.5-3b & 9 & 6 & 7 & 2 & 96 & 2 & 15 & 63 \\ \hline
Gemma2-2b & 9 & 0 & 0 & 46 & 100 & 1 & 44 & 0 \\ \hline
Granite3.2-2b & 0 & 0 & 0 & 51 & 100 & 0 & 41 & 8 \\ \hline
Granite3.3-2b & 0 & 0 & 0 & 0 & 100 & 0 & 0 & 100 \\ \hline

\end{tabular} }
\end{table*}

\begin{table*}[ht]
\centering
\caption{Static Role Self-Alignment on TriviaQA}
\label{sl9}
\resizebox{1\textwidth}{!}{
\begin{tabular}{lrrrrrrrr}
\hline
Model & Mathematician & Physicist & Physician & Scientist & Literacy Specialist & Engineer & Software Developer & Polymath \\
\hline
Phi4-14b & 47 & 48 & 1 & 24 & 27 & 1 & 9 & 43 \\ \hline
Qwen2.5-14b & 47 & 50 & 0 & 8 & 43 & 2 & 1 & 49 \\ \hline
Gemma3-12b & 0 & 0 & 0 & 0 & 0 & 0 & 100 & 100 \\ \hline
Mistral-nemo-12b & 0 & 1 & 0 & 80 & 39 & 4 & 9 & 67 \\ \hline
Gemma2-9b & 1 & 0 & 0 & 80 & 17 & 0 & 7 & 94 \\ \hline
Granite3.2-8b & 4 & 27 & 2 & 92 & 54 & 2 & 9 & 10 \\ \hline
Granite3.3-8b & 2 & 9 & 1 & 61 & 70 & 1 & 55 & 1 \\ \hline
Llama3.1-8b & 10 & 80 & 2 & 65 & 11 & 1 & 1 & 30 \\ \hline
Mistral-7b & 49 & 70 & 3 & 31 & 2 & 21 & 6 & 2 \\ \hline
Qwen2.5-7b & 50 & 55 & 3 & 30 & 21 & 1 & 1 & 39 \\ \hline
Llama3.2-3b & 19 & 24 & 6 & 66 & 29 & 3 & 0 & 52 \\ \hline
Qwen2.5-3b & 42 & 38 & 1 & 28 & 52 & 5 & 10 & 23 \\ \hline
Gemma2-2b & 27 & 8 & 9 & 77 & 42 & 3 & 6 & 28 \\ \hline
Granite3.2-2b & 62 & 71 & 5 & 25 & 3 & 13 & 4 & 16 \\ \hline
Granite3.3-2b & 9 & 1 & 13 & 40 & 77 & 34 & 12 & 14 \\ \hline
\end{tabular} }
\end{table*}

\subsection{Dynamic Self-Alignment}
 Tables \ref{role_self_alignment_part1} and \ref{role_self_alignment_part2} shows the role selected by each model in dynamic self-Alignment.
\begin{table*}[ht]
\centering
\caption{Dynamic Role Self-Alignment across three reasoning benchmarks (AQuA, GSM8K, and OpenbookQA). The numbers in parentheses indicate the total occurrences of the role among 100 model responses.}
\label{role_self_alignment_part1}
 \centering
\resizebox{\textwidth}{!}{
\begin{tabular}{lrrr}
\hline
Model & AQuA & GSM8K & OpenbookQA \\
\hline
GPT-4.1          & Mathematics Educator (58) & Mathematician (87) & Cognitive Scientist (76) \\
GPT-4.1-mini     & Mathematics Educator (95) & Mathematician (89) & Cognitive Scientist (81) \\
GPT-4.1-nano     & Mathematics Educator (89) & Mathematics Educator (40) & Science Educator (37) \\
GPT-4o           & Mathematician (61) & Mathematician (99) & Cognitive Scientist (96) \\
GPT-4o-mini      & Mathematics Educator (38) & Mathematician (75) & Research Scientist (43) \\
Phi4-14b         & Mathematics Educator (31) & Mathematics Educator (32) & Cognitive Scientist (14) \\
Qwen2.5-14b      & Mathematics Educator (75) & Math Teacher (54) & Knowledge Engineer (85) \\
\hline
\end{tabular}}
\end{table*}

\begin{table*}[ht]
\centering
\caption{Dynamic Role Self-Alignment across three reasoning benchmarks (StrategyQA, RACE(M\&h), and TriviaQA). The numbers in parentheses indicate the total occurrences of the role among 100 model responses.}
\label{role_self_alignment_part2}
 \centering
\resizebox{\textwidth}{!}{
\begin{tabular}{lrrr}
\hline
Model & StrategyQA & RACE(M\&h) & TriviaQA \\
\hline
GPT-4.1          & Research Scientist (39) & English Teacher (81) & Trivia Expert (62) \\
GPT-4.1-mini     & Data Scientist (74) & Reading Comprehension Specialist (48) & Data Scientist (68) \\
GPT-4.1-nano     & Logic Expert (21) & Educational Researcher (24) & Reading Comprehension Researcher (28) \\
GPT-4o           & Cognitive Scientist (82) & English Teacher (26) & Trivia Expert (63) \\
GPT-4o-mini      & Data Scientist (75) & Educational Assessment Specialist (43) & Data Scientist (98) \\
Phi4-14b         & Ai Research Scientist (13) & Reading Specialist (13) & Knowledge Engineer (22) \\
Qwen2.5-14b      & Strategist (24) & Educational Assessment Specialist (44) & Information Scientist (65) \\
\hline
\end{tabular}}
\end{table*}

\section{Detailed Results}
\label{Sec:E}
This section presents the comprehensive results, encompassing all versions, variants, and the respective roles displayed for each model. These results are detailed in Tables~\ref{GPT-4.1} to \ref{dynamic_Granite332b}.
\begin{table}[!t]
  \caption{Results for Static Role-Alignment on all prompt variants of GPT-4.1 on AQuA-RAT, GSM8K, OpenbookQA, StrategyQA, RACE-h, RACE-m, TriviaQA. Symbol 'X' indicates that the role was not in the selected roles.}
  \label{GPT-4.1}
  \centering
  \resizebox{0.8\linewidth}{!}{%
  \begin{tabular}{lccccccc}
    \toprule
    \textbf{Prompt} & \textbf{AQuA-RAT} & \textbf{GSM8K} & \textbf{OpenbookQA} & \textbf{StrategyQA} & \textbf{RACE-h} & \textbf{RACE-m} & \textbf{TriviaQA} \\
    \midrule
    Mathematician (Pre-question) & X & X & X & 84.41 & X & X & X \\
    \midrule
    Mathematician (Post-question) & X & X & X & 84.24 & X & X & X \\
    \midrule
    Mathematician (Split) & X & X & X & 84.89 & X & X & X \\
    \midrule
    Physicist (Pre-question) & X & X & X & X & X & X & X \\
    \midrule
    Physicist (Post-question) & X & X & X & X & X & X & X \\
    \midrule
    Physicist (Split) & X & X & X & X & X & X & X \\
    \midrule
    Physician (Pre-question) & X & X & X & X & X & 95.19 & X \\
\bottomrule
  \end{tabular}%
  }
\end{table}

\begin{table}[!t]
  \caption{Results for Static Role-Alignment on all prompt variants of GPT-4.1-mini on AQuA-RAT, GSM8K, OpenbookQA, StrategyQA, RACE-h, RACE-m, TriviaQA.}
  \label{GPT-4.1-mini}
  \centering
  \resizebox{0.8\linewidth}{!}{%
  \begin{tabular}{lccccccc}
    \toprule
    \textbf{Prompt} & \textbf{AQuA-RAT} & \textbf{GSM8K} & \textbf{OpenbookQA} & \textbf{StrategyQA} & \textbf{RACE-h} & \textbf{RACE-m} & \textbf{TriviaQA} \\
    \midrule
    Mathematician (Pre-question) & 79.53 & 87.64 & 92.60 & 82.27 & 89.05 & 92.97 & 84.20 \\
    \midrule
    Mathematician (Post-question) & 81.50 & 86.43 & 94.40 & 81.88 & 89.51 & 94.08 & 83.90 \\
    \midrule
    Mathematician (Split) & 78.74 & 86.13 & 95.60 & 81.75 & 89.71 & 93.87 & 84.50 \\
    \midrule
    Physicist (Pre-question) & 81.10 & 87.49 & 95.40 & 81.44 & 89.91 & 93.38 & 83.70 \\
    \midrule
    Physicist (Post-question) & 77.95 & 85.75 & 93.80 & 81.35 & 89.19 & 93.87 & 83.90 \\
    \midrule
    Physicist (Split) & 79.53 & 86.13 & 94.40 & 81.40 & 90.14 & 93.11 & 83.40 \\
    \midrule
    Physician (Pre-question) & 85.43 & 86.43 & 95.20 & 80.92 & 90.37 & 93.45 & 82.50 \\
\bottomrule
  \end{tabular}%
  }
\end{table}

\begin{table}[!t]
  \caption{Results for Static Role-Alignment on all prompt variants of GPT-4.1-nano on AQuA-RAT, GSM8K, OpenbookQA, StrategyQA, RACE-h, RACE-m, TriviaQA.}
  \label{GPT-4.1-nano}
  \centering
  \resizebox{0.8\linewidth}{!}{%
  \begin{tabular}{lccccccc}
    \toprule
    \textbf{Prompt} & \textbf{AQuA-RAT} & \textbf{GSM8K} & \textbf{OpenbookQA} & \textbf{StrategyQA} & \textbf{RACE-h} & \textbf{RACE-m} & \textbf{TriviaQA} \\
    \midrule
    Mathematician (Pre-question) & 65.75 & 86.66 & 87.20 & 76.03 & 79.82 & 87.19 & 76.60 \\
    \midrule
    Mathematician (Post-question) & 65.75 & 84.53 & 86.20 & 75.11 & 82.33 & 89.42 & 79.50 \\
    \midrule
    Mathematician (Split) & 62.99 & 84.84 & 86.20 & 75.28 & 81.59 & 89.97 & 78.70 \\
    \midrule
    Physicist (Pre-question) & 67.32 & 85.60 & 88.40 & 72.84 & 79.65 & 87.60 & 76.40 \\
    \midrule
    Physicist (Post-question) & 61.02 & 83.24 & 87.60 & 73.71 & 81.65 & 90.04 & 78.50 \\
    \midrule
    Physicist (Split) & 62.99 & 83.24 & 86.00 & 73.80 & 81.22 & 89.35 & 79.30 \\
    \midrule
    Physician (Pre-question) & 77.95 & 85.44 & 88.20 & 68.03 & 78.04 & 87.40 & 76.60 \\
\bottomrule
  \end{tabular}%
  }
\end{table}

\begin{table}[!t]
  \caption{Results for Static Role-Alignment on all prompt variants of GPT-4o on AQuA-RAT, GSM8K, OpenbookQA, StrategyQA, RACE-h, RACE-m, TriviaQA. Symbol 'X' indicates that the role was not in the selected roles.}
  \label{GPT-4o}
  \centering
  \resizebox{0.8\linewidth}{!}{%
  \begin{tabular}{lccccccc}
    \toprule
    \textbf{Prompt} & \textbf{AQuA-RAT} & \textbf{GSM8K} & \textbf{OpenbookQA} & \textbf{StrategyQA} & \textbf{RACE-h} & \textbf{RACE-m} & \textbf{TriviaQA} \\
    \midrule
    Mathematician (Pre-question) & 64.57 & 88.48 & 94.80 & 80.22 & 7.15 & 94.22 & 85.70 \\
    \midrule
    Mathematician (Post-question) & 68.90 & 80.82 & 91.00 & 82.45 & X & 92.13 & 85.80 \\
    \midrule
    Mathematician (Split) & 70.08 & X & 94.60 & 81.22 & X & 7.94 & 85.70 \\
    \midrule
    Physicist (Pre-question) & 80.31 & X & 91.80 & X & X & X & 72.10 \\
    \midrule
    Physicist (Post-question) & 72.44 & X & 86.40 & X & X & X & X \\
    \midrule
    Physicist (Split) & 74.41 & X & 89.20 & X & X & X & X \\
    \midrule
    Physician (Pre-question) & 79.13 & X & 93.40 & X & X & 94.08 & X \\
\bottomrule
  \end{tabular}%
  }
\end{table}

\begin{table}[!t]
  \caption{Results for Static Role-Alignment on all prompt variants of GPT-4o-mini on AQuA-RAT, GSM8K, OpenbookQA, StrategyQA, RACE-h, RACE-m, TriviaQA.}
  \label{GPT-4o-mini}
  \centering
  \resizebox{0.8\linewidth}{!}{%
  \begin{tabular}{lccccccc}
    \toprule
    \textbf{Prompt} & \textbf{AQuA-RAT} & \textbf{GSM8K} & \textbf{OpenbookQA} & \textbf{StrategyQA} & \textbf{RACE-h} & \textbf{RACE-m} & \textbf{TriviaQA} \\
    \midrule
    Mathematician (Pre-question) & 68.50 & 88.32 & 82.40 & 78.82 & 86.62 & 92.76 & 78.80 \\
    \midrule
    Mathematician (Post-question) & 62.60 & 88.70 & 75.60 & 79.30 & 86.99 & 92.20 & 79.10 \\
    \midrule
    Mathematician (Split) & 66.93 & 87.72 & 81.00 & 77.99 & 86.19 & 92.48 & 78.40 \\
    \midrule
    Physicist (Pre-question) & 72.05 & 87.49 & 83.80 & 78.30 & 87.31 & 92.83 & 78.30 \\
    \midrule
    Physicist (Post-question) & 69.69 & 88.70 & 74.40 & 78.34 & 86.54 & 92.20 & 79.00 \\
    \midrule
    Physicist (Split) & 67.72 & 87.34 & 78.20 & 77.90 & 86.74 & 92.20 & 79.20 \\
    \midrule
    Physician (Pre-question) & 75.98 & 87.95 & 83.80 & 77.25 & 86.59 & 92.83 & 78.40 \\
\bottomrule
  \end{tabular}%
  }
\end{table}

\begin{table}[!t]
  \caption{Results for Static Role-Alignment on all prompt variants of Phi4-14b on AQuA-RAT, GSM8K, OpenbookQA, StrategyQA, RACE-h, RACE-m, TriviaQA. Symbol 'X' indicates that the role was not in the top two selected roles.}
  \label{Phi4-14b}
  \centering
  \resizebox{0.8\linewidth}{!}{%
  \begin{tabular}{lccccccc}
    \toprule
    \textbf{Prompt} & \textbf{AQuA-RAT} & \textbf{GSM8K} & \textbf{OpenbookQA} & \textbf{StrategyQA} & \textbf{RACE-h} & \textbf{RACE-m} & \textbf{TriviaQA} \\
    \midrule
    Mathematician (Pre-question) & 52.36 & 89.54 & X & X & X & X & 74.60 \\
    \midrule
    Mathematician (Post-question) & 66.54 & 87.57 & X & X & X & X & 73.90 \\
    \midrule
    Mathematician (Split) & 54.72 & 88.70 & X & X & X & X & 73.90 \\
    \midrule
    Physicist (Pre-question) & X & X & X & X & X & X & 74.40 \\
    \midrule
    Physicist (Post-question) & X & X & X & X & X & X & 73.90 \\
    \midrule
    Physicist (Split) & X & X & X & X & X & X & 72.90 \\
    \midrule
    Physician (Pre-question) & X & X & X & X & X & X & X \\
\bottomrule
  \end{tabular}%
  }
\end{table}

\begin{table}[!t]
  \caption{Results for Static Role-Alignment on all prompt variants of Qwen2.5-14b on AQuA-RAT, GSM8K, OpenbookQA, StrategyQA, RACE-h, RACE-m, TriviaQA. Symbol 'X' indicates that the role was not in the top two selected roles.}
  \label{Qwen2.5-14b}
  \centering
  \resizebox{0.8\linewidth}{!}{%
  \begin{tabular}{lccccccc}
    \toprule
    \textbf{Prompt} & \textbf{AQuA-RAT} & \textbf{GSM8K} & \textbf{OpenbookQA} & \textbf{StrategyQA} & \textbf{RACE-h} & \textbf{RACE-m} & \textbf{TriviaQA} \\
    \midrule
    Mathematician (Pre-question) & 59.06 & 87.34 & X & X & X & X & X \\
    \midrule
    Mathematician (Post-question) & 59.06 & 87.26 & X & X & X & X & X \\
    \midrule
    Mathematician (Split) & 52.76 & 85.97 & X & X & X & X & X \\
    \midrule
    Physicist (Pre-question) & 64.57 & 0.61 & X & X & X & X & 72.90 \\
    \midrule
    Physicist (Post-question) & 64.57 & X & X & X & X & X & 72.60 \\
    \midrule
    Physicist (Split) & 61.81 & X & X & X & X & X & 72.30 \\
    \midrule
    Physician (Pre-question) & X & X & X & X & X & X & X \\
\bottomrule
  \end{tabular}%
  }
\end{table}

\begin{table}[!t]
  \caption{Results for Static Role-Alignment on all prompt variants of Gemma3-12b on AQuA-RAT, GSM8K, OpenbookQA, StrategyQA, RACE-h, RACE-m, TriviaQA. Symbol 'X' indicates that the role was not in the top two selected roles.}
  \label{Gemma3-12b}
  \centering
  \resizebox{0.8\linewidth}{!}{%
  \begin{tabular}{lccccccc}
    \toprule
    \textbf{Prompt} & \textbf{AQuA-RAT} & \textbf{GSM8K} & \textbf{OpenbookQA} & \textbf{StrategyQA} & \textbf{RACE-h} & \textbf{RACE-m} & \textbf{TriviaQA} \\
    \midrule
    Mathematician (Pre-question) & X & 86.58 & X & X & X & X & X \\
    \midrule
    Mathematician (Post-question) & X & 84.23 & X & X & X & X & X \\
    \midrule
    Mathematician (Split) & X & 84.08 & X & X & X & X & X \\
    \midrule
    Physicist (Pre-question) & X & X & X & X & X & X & X \\
    \midrule
    Physicist (Post-question) & X & X & X & X & X & X & X \\
    \midrule
    Physicist (Split) & X & X & X & X & X & X & X \\
    \midrule
    Physician (Pre-question) & X & X & X & X & X & X & X \\
\bottomrule
  \end{tabular}%
  }
\end{table}

\begin{table}[!t]
  \caption{Results for Static Role-Alignment on all prompt variants of Mistral-nemo-12b on AQuA-RAT, GSM8K, OpenbookQA, StrategyQA, RACE-h, RACE-m, TriviaQA. Symbol 'X' indicates that the role was not in the top two selected roles.}
  \label{Mistral-nemo-12b}
  \centering
  \resizebox{0.8\linewidth}{!}{%
  \begin{tabular}{lccccccc}
    \toprule
    \textbf{Prompt} & \textbf{AQuA-RAT} & \textbf{GSM8K} & \textbf{OpenbookQA} & \textbf{StrategyQA} & \textbf{RACE-h} & \textbf{RACE-m} & \textbf{TriviaQA} \\
    \midrule
    Mathematician (Pre-question) & 58.66 & 80.06 & X & X & X & X & X \\
    \midrule
    Mathematician (Post-question) & 57.09 & 79.45 & X & X & X & X & X \\
    \midrule
    Mathematician (Split) & 59.45 & 80.82 & X & X & X & X & X \\
    \midrule
    Physicist (Pre-question) & X & X & X & X & X & X & X \\
    \midrule
    Physicist (Post-question) & X & X & X & X & X & X & X \\
    \midrule
    Physicist (Split) & X & X & X & X & X & X & X \\
    \midrule
    Physician (Pre-question) & X & X & X & X & X & X & X \\
\bottomrule
  \end{tabular}%
  }
\end{table}

\begin{table}[!t]
  \caption{Results for Static Role-Alignment on all prompt variants of Gemma2-9b on AQuA-RAT, GSM8K, OpenbookQA, StrategyQA, RACE-h, RACE-m, TriviaQA. Symbol 'X' indicates that the role was not in the top two selected roles.}
  \label{Gemma2-9b}
  \centering
  \resizebox{0.8\linewidth}{!}{%
  \begin{tabular}{lccccccc}
    \toprule
    \textbf{Prompt} & \textbf{AQuA-RAT} & \textbf{GSM8K} & \textbf{OpenbookQA} & \textbf{StrategyQA} & \textbf{RACE-h} & \textbf{RACE-m} & \textbf{TriviaQA} \\
    \midrule
    Mathematician (Pre-question) & 55.51 & 84.00 & X & X & X & X & X \\
    \midrule
    Mathematician (Post-question) & 53.54 & 84.84 & X & X & X & X & X \\
    \midrule
    Mathematician (Split) & 52.36 & 84.38 & X & X & X & X & X \\
    \midrule
    Physicist (Pre-question) & X & X & X & X & X & X & X \\
    \midrule
    Physicist (Post-question) & X & X & X & X & X & X & X \\
    \midrule
    Physicist (Split) & X & X & X & X & X & X & X \\
    \midrule
    Physician (Pre-question) & X & X & X & X & X & X & X \\
\bottomrule
  \end{tabular}%
  }
\end{table}

\begin{table}[!t]
  \caption{Results for Static Role-Alignment on all prompt variants of Granite3.2-8b on AQuA-RAT, GSM8K, OpenbookQA, StrategyQA, RACE-h, RACE-m, TriviaQA. Symbol 'X' indicates that the role was not in the top two selected roles.}
  \label{Granite3.2-8b}
  \centering
  \resizebox{0.8\linewidth}{!}{%
  \begin{tabular}{lccccccc}
    \toprule
    \textbf{Prompt} & \textbf{AQuA-RAT} & \textbf{GSM8K} & \textbf{OpenbookQA} & \textbf{StrategyQA} & \textbf{RACE-h} & \textbf{RACE-m} & \textbf{TriviaQA} \\
    \midrule
    Mathematician (Pre-question) & 54.72 & 77.10 & X & X & X & X & X \\
    \midrule
    Mathematician (Post-question) & 57.48 & 79.38 & X & X & X & X & X \\
    \midrule
    Mathematician (Split) & 60.63 & 76.88 & X & X & X & X & X \\
    \midrule
    Physicist (Pre-question) & X & X & 45.80 & 57.82 & X & X & X \\
    \midrule
    Physicist (Post-question) & X & X & 63.60 & 59.04 & X & X & X \\
    \midrule
    Physicist (Split) & X & X & 66.60 & 59.34 & X & X & X \\
    \midrule
    Physician (Pre-question) & X & X & X & X & X & X & X \\
\bottomrule
  \end{tabular}%
  }
\end{table}

\begin{table}[!t]
  \caption{Results for Static Role-Alignment on all prompt variants of Granite3.3-8b on AQuA-RAT, GSM8K, OpenbookQA, StrategyQA, RACE-h, RACE-m, TriviaQA. Symbol 'X' indicates that the role was not in the top two selected roles.}
  \label{Granite3.3-8b}
  \centering
  \resizebox{0.8\linewidth}{!}{%
  \begin{tabular}{lccccccc}
    \toprule
    \textbf{Prompt} & \textbf{AQuA-RAT} & \textbf{GSM8K} & \textbf{OpenbookQA} & \textbf{StrategyQA} & \textbf{RACE-h} & \textbf{RACE-m} & \textbf{TriviaQA} \\
    \midrule
    Mathematician (Pre-question) & 62.60 & 77.94 & 70.60 & 61.92 & X & X & X \\
    \midrule
    Mathematician (Post-question) & 57.09 & 78.77 & 75.40 & 67.12 & X & X & X \\
    \midrule
    Mathematician (Split) & 59.06 & 79.30 & 71.40 & 64.98 & X & X & X \\
    \midrule
    Physicist (Pre-question) & X & X & X & X & X & X & X \\
    \midrule
    Physicist (Post-question) & X & X & X & X & X & X & X \\
    \midrule
    Physicist (Split) & X & X & X & X & X & X & X \\
    \midrule
    Physician (Pre-question) & X & X & X & X & X & X & X \\
\bottomrule
  \end{tabular}%
  }
\end{table}

\begin{table}[!t]
  \caption{Results for Static Role-Alignment on all prompt variants of Llama3.1-8b on AQuA-RAT, GSM8K, OpenbookQA, StrategyQA, RACE-h, RACE-m, TriviaQA. Symbol 'X' indicates that the role was not in the top two selected roles.}
  \label{Llama3.1-8b}
  \centering
  \resizebox{0.8\linewidth}{!}{%
  \begin{tabular}{lccccccc}
    \toprule
    \textbf{Prompt} & \textbf{AQuA-RAT} & \textbf{GSM8K} & \textbf{OpenbookQA} & \textbf{StrategyQA} & \textbf{RACE-h} & \textbf{RACE-m} & \textbf{TriviaQA} \\
    \midrule
    Mathematician (Pre-question) & 60.24 & 79.45 & X & X & X & X & X \\
    \midrule
    Mathematician (Post-question) & 54.72 & 79.38 & X & X & X & X & X \\
    \midrule
    Mathematician (Split) & 59.06 & 79.53 & X & X & X & X & X \\
    \midrule
    Physicist (Pre-question) & X & X & 51.80 & 67.99 & X & X & 71.10 \\
    \midrule
    Physicist (Post-question) & X & X & 39.00 & 64.54 & X & X & 73.00 \\
    \midrule
    Physicist (Split) & X & X & 51.80 & 69.04 & X & X & 72.90 \\
    \midrule
    Physician (Pre-question) & X & X & X & X & X & X & X \\
\bottomrule
  \end{tabular}%
  }
\end{table}

\begin{table}[!t]
  \caption{Results for Static Role-Alignment on all prompt variants of Mistral-7b on AQuA-RAT, GSM8K, OpenbookQA, StrategyQA, RACE-h, RACE-m, TriviaQA. Symbol 'X' indicates that the role was not in the top two selected roles.}
  \label{Mistral-7b}
  \centering
  \resizebox{0.8\linewidth}{!}{%
  \begin{tabular}{lccccccc}
    \toprule
    \textbf{Prompt} & \textbf{AQuA-RAT} & \textbf{GSM8K} & \textbf{OpenbookQA} & \textbf{StrategyQA} & \textbf{RACE-h} & \textbf{RACE-m} & \textbf{TriviaQA} \\
    \midrule
    Mathematician (Pre-question) & 14.96 & 42.84 & X & 35.07 & X & X & 45.50 \\
    \midrule
    Mathematician (Post-question) & 15.35 & 41.02 & X & 34.10 & X & X & 46.90 \\
    \midrule
    Mathematician (Split) & 20.08 & 41.77 & X & 36.11 & X & X & 46.00 \\
    \midrule
    Physicist (Pre-question) & 17.72 & 42.53 & 35.20 & 33.36 & 24.70 & 39.69 & 42.20 \\
    \midrule
    Physicist (Post-question) & 12.60 & 42.23 & 33.80 & 26.24 & 16.92 & 30.29 & 40.30 \\
    \midrule
    Physicist (Split) & 16.14 & 40.49 & 34.20 & 29.21 & 19.18 & 33.29 & 42.70 \\
    \midrule
    Physician (Pre-question) & X & X & X & X & X & X & X \\
\bottomrule
  \end{tabular}%
  }
\end{table}

\begin{table}[!t]
  \caption{Results for Static Role-Alignment on all prompt variants of Qwen2.5-7b on AQuA-RAT, GSM8K, OpenbookQA, StrategyQA, RACE-h, RACE-m, TriviaQA. Symbol 'X' indicates that the role was not in the top two selected roles.}
  \label{Qwen2.5-7b}
  \centering
  \resizebox{0.8\linewidth}{!}{%
  \begin{tabular}{lccccccc}
    \toprule
    \textbf{Prompt} & \textbf{AQuA-RAT} & \textbf{GSM8K} & \textbf{OpenbookQA} & \textbf{StrategyQA} & \textbf{RACE-h} & \textbf{RACE-m} & \textbf{TriviaQA} \\
    \midrule
    Mathematician (Pre-question) & 61.81 & 82.79 & X & X & X & X & 64.60 \\
    \midrule
    Mathematician (Post-question) & 66.14 & 84.69 & X & X & X & X & 63.90 \\
    \midrule
    Mathematician (Split) & 70.87 & 85.97 & X & X & X & X & 63.90 \\
    \midrule
    Physicist (Pre-question) & X & X & X & X & X & X & 63.60 \\
    \midrule
    Physicist (Post-question) & X & X & X & X & X & X & 64.50 \\
    \midrule
    Physicist (Split) & X & X & X & X & X & X & 63.40 \\
    \midrule
    Physician (Pre-question) & X & X & X & X & X & X & X \\
\bottomrule
  \end{tabular}%
  }
\end{table}

\begin{table}[!t]
  \caption{Results for Static Role-Alignment on all prompt variants of Llama3.2-3b on AQuA-RAT, GSM8K, OpenbookQA, StrategyQA, RACE-h, RACE-m, TriviaQA. Symbol 'X' indicates that the role was not in the top two selected roles.}
  \label{Llama3.2-3b}
  \centering
  \resizebox{0.8\linewidth}{!}{%
  \begin{tabular}{lccccccc}
    \toprule
    \textbf{Prompt} & \textbf{AQuA-RAT} & \textbf{GSM8K} & \textbf{OpenbookQA} & \textbf{StrategyQA} & \textbf{RACE-h} & \textbf{RACE-m} & \textbf{TriviaQA} \\
    \midrule
    Mathematician (Pre-question) & 51.57 & X & X & X & X & X & X \\
    \midrule
    Mathematician (Post-question) & 51.18 & X & X & X & X & X & X \\
    \midrule
    Mathematician (Split) & 51.97 & X & X & X & X & X & X \\
    \midrule
    Physicist (Pre-question) & X & X & 57.80 & 66.90 & X & X & X \\
    \midrule
    Physicist (Post-question) & X & X & 55.20 & 64.59 & X & X & X \\
    \midrule
    Physicist (Split) & X & X & 57.80 & 64.06 & X & X & X \\
    \midrule
    Physician (Pre-question) & X & X & X & X & X & X & X \\
\bottomrule
  \end{tabular}%
  }
\end{table}

\begin{table}[!t]
  \caption{Results for Static Role-Alignment on all prompt variants of Qwen2.5-3b on AQuA-RAT, GSM8K, OpenbookQA, StrategyQA, RACE-h, RACE-m, TriviaQA. Symbol 'X' indicates that the role was not in the top two selected roles.}
  \label{Qwen2.5-3b}
  \centering
  \resizebox{0.8\linewidth}{!}{%
  \begin{tabular}{lccccccc}
    \toprule
    \textbf{Prompt} & \textbf{AQuA-RAT} & \textbf{GSM8K} & \textbf{OpenbookQA} & \textbf{StrategyQA} & \textbf{RACE-h} & \textbf{RACE-m} & \textbf{TriviaQA} \\
    \midrule
    Mathematician (Pre-question) & 58.27 & 72.40 & 52.60 & 57.86 & X & X & 55.30 \\
    \midrule
    Mathematician (Post-question) & 58.66 & 76.72 & 54.00 & 60.92 & X & X & 55.20 \\
    \midrule
    Mathematician (Split) & 59.45 & 78.17 & 55.60 & 60.39 & X & X & 54.90 \\
    \midrule
    Physicist (Pre-question) & X & X & 48.60 & 57.51 & X & X & X \\
    \midrule
    Physicist (Post-question) & X & X & 51.20 & 59.87 & X & X & X \\
    \midrule
    Physicist (Split) & X & X & 49.40 & 59.91 & X & X & X \\
    \midrule
    Physician (Pre-question) & X & X & X & X & X & X & X \\
\bottomrule
  \end{tabular}%
  }
\end{table}

\begin{table}[!t]
  \caption{Results for Static Role-Alignment on all prompt variants of Gemma2-2b on AQuA-RAT, GSM8K, OpenbookQA, StrategyQA, RACE-h, RACE-m, TriviaQA. Symbol 'X' indicates that the role was not in the top two selected roles.}
  \label{Gemma2-2b}
  \centering
  \resizebox{0.8\linewidth}{!}{%
  \begin{tabular}{lccccccc}
    \toprule
    \textbf{Prompt} & \textbf{AQuA-RAT} & \textbf{GSM8K} & \textbf{OpenbookQA} & \textbf{StrategyQA} & \textbf{RACE-h} & \textbf{RACE-m} & \textbf{TriviaQA} \\
    \midrule
    Mathematician (Pre-question) & 28.35 & 62.47 & X & 57.12 & X & X & X \\
    \midrule
    Mathematician (Post-question) & 26.38 & 61.56 & X & 54.98 & X & X & X \\
    \midrule
    Mathematician (Split) & 25.59 & 61.41 & X & 54.45 & X & X & X \\
    \midrule
    Physicist (Pre-question) & X & X & X & X & X & X & X \\
    \midrule
    Physicist (Post-question) & X & X & X & X & X & X & X \\
    \midrule
    Physicist (Split) & X & X & X & X & X & X & X \\
    \midrule
    Physician (Pre-question) & X & X & X & X & X & X & X \\
\bottomrule
  \end{tabular}%
  }
\end{table}

\begin{table}[!t]
  \caption{Results for Static Role-Alignment on all prompt variants of Granite3.2-2b on AQuA-RAT, GSM8K, OpenbookQA, StrategyQA, RACE-h, RACE-m, TriviaQA. Symbol 'X' indicates that the role was not in the top two selected roles.}
  \label{Granite3.2-2b}
  \centering
  \resizebox{0.8\linewidth}{!}{%
  \begin{tabular}{lccccccc}
    \toprule
    \textbf{Prompt} & \textbf{AQuA-RAT} & \textbf{GSM8K} & \textbf{OpenbookQA} & \textbf{StrategyQA} & \textbf{RACE-h} & \textbf{RACE-m} & \textbf{TriviaQA} \\
    \midrule
    Mathematician (Pre-question) & 41.34 & 63.15 & 50.80 & 53.58 & 54.55 & 59.89 & 52.80 \\
    \midrule
    Mathematician (Post-question) & 41.73 & 62.09 & 50.40 & 54.06 & 55.60 & 60.58 & 52.80 \\
    \midrule
    Mathematician (Split) & 40.16 & 63.68 & 50.80 & 56.46 & 56.09 & 63.51 & 53.00 \\
    \midrule
    Physicist (Pre-question) & 42.91 & 61.18 & 50.20 & 53.45 & 56.75 & 62.47 & 54.30 \\
    \midrule
    Physicist (Post-question) & 39.37 & 64.14 & 49.80 & 53.97 & 57.72 & 62.88 & 54.70 \\
    \midrule
    Physicist (Split) & 39.37 & 62.77 & 52.00 & 55.20 & 56.83 & 61.56 & 53.50 \\
    \midrule
    Physician (Pre-question) & X & X & X & X & X & X & X \\
\bottomrule
  \end{tabular}%
  }
\end{table}

\begin{table}[!t]
  \caption{Results for Static Role-Alignment on all prompt variants of Granite3.3-2b on AQuA-RAT, GSM8K, OpenbookQA, StrategyQA, RACE-h, RACE-m, TriviaQA. Symbol 'X' indicates that the role was not in the top two selected roles.}
  \label{Granite3.3-2b}
  \centering
  \resizebox{0.8\linewidth}{!}{%
  \begin{tabular}{lccccccc}
    \toprule
    \textbf{Prompt} & \textbf{AQuA-RAT} & \textbf{GSM8K} & \textbf{OpenbookQA} & \textbf{StrategyQA} & \textbf{RACE-h} & \textbf{RACE-m} & \textbf{TriviaQA} \\
    \midrule
    Mathematician (Pre-question) & 53.15 & 70.43 & X & 55.24 & X & X & X \\
    \midrule
    Mathematician (Post-question) & 51.57 & 73.62 & X & 58.08 & X & X & X \\
    \midrule
    Mathematician (Split) & 53.15 & 69.14 & X & 56.81 & X & X & X \\
    \midrule
    Physicist (Pre-question) & X & X & X & X & X & X & X \\
    \midrule
    Physicist (Post-question) & X & X & X & X & X & X & X \\
    \midrule
    Physicist (Split) & X & X & X & X & X & X & X \\
    \midrule
    Physician (Pre-question) & X & X & X & X & X & X & X \\
\bottomrule
  \end{tabular}%
  }
\end{table}

\begin{table}[!t]
  \caption{Performance of dynamically-selected roles for GPT-4.1.}
  \label{dynamic_GPT41}
  \centering
 \resizebox{0.8\linewidth}{!}{ 
  \begin{tabular}{lc}
    \toprule
    \textbf{Dataset and Prompt Variant} & \textbf{Accuracy} \\
    \midrule
    \multicolumn{2}{l}{\textit{AQuA-RAT}} \\
    Mathematics Educator (Pre-question) & 68.50 \\
    Mathematics Educator (Post-question) & 62.20 \\
    Mathematics Educator (Split) & 61.81 \\
    \midrule
    \multicolumn{2}{l}{\textit{GSM8K}} \\
    (Same as Static) & $\equiv$ \\
    \midrule
    \multicolumn{2}{l}{\textit{OpenbookQA}} \\
    Cognitive Scientist (Pre-question) & 93.00 \\
    Cognitive Scientist (Post-question) & 85.80 \\
    Cognitive Scientist (Split) & 82.80 \\
    \midrule
    \multicolumn{2}{l}{\textit{StrategyQA}} \\
    Research Scientist (Pre-question) & 84.54 \\
    Research Scientist (Post-question) & 83.89 \\
    Research Scientist (Split) & 84.45 \\
    \midrule
    \multicolumn{2}{l}{\textit{RACE-h}} \\
    English Teacher (Pre-question) & 92.08 \\
    English Teacher (Post-question) & 91.94 \\
    English Teacher (Split) & 91.85 \\
    \midrule
    \multicolumn{2}{l}{\textit{RACE-m}} \\
    English Teacher (Pre-question) & 95.33 \\
    English Teacher (Post-question) & 95.13 \\
    English Teacher (Split) & 95.47 \\
    \midrule
    \multicolumn{2}{l}{\textit{TriviaQA}} \\
    Trivia Expert (Pre-question) & 86.90 \\
    Trivia Expert (Post-question) & 87.10 \\
    Trivia Expert (Split) & 87.50 \\
    \midrule
    \bottomrule
  \end{tabular}
} 
\end{table}

\begin{table}[!t]
  \caption{Performance of dynamically-selected roles for GPT-4.1-mini.}
  \label{dynamic_GPT41mini}
  \centering
 \resizebox{0.8\linewidth}{!}{ 
  \begin{tabular}{lc}
    \toprule
    \textbf{Dataset and Prompt Variant} & \textbf{Accuracy} \\
    \midrule
    \multicolumn{2}{l}{\textit{AQuA-RAT}} \\
    Mathematics Educator (Pre-question) & 74.80 \\
    Mathematics Educator (Post-question) & 78.35 \\
    Mathematics Educator (Split) & 77.17 \\
    \midrule
    \multicolumn{2}{l}{\textit{GSM8K}} \\
    Mathematician (Pre-question) & 87.11 \\
    Mathematician (Post-question) & 86.35 \\
    Mathematician (Split) & 86.35 \\
    \midrule
    \multicolumn{2}{l}{\textit{OpenbookQA}} \\
    Cognitive Scientist (Pre-question) & 94.60 \\
    Cognitive Scientist (Post-question) & 95.40 \\
    Cognitive Scientist (Split) & 96.40 \\
    \midrule
    \multicolumn{2}{l}{\textit{StrategyQA}} \\
    Data Scientist (Pre-question) & 81.18 \\
    Data Scientist (Post-question) & 81.31 \\
    Data Scientist (Split) & 81.66 \\
    \midrule
    \multicolumn{2}{l}{\textit{RACE-h}} \\
    Reading Comprehension Specialist (Pre-question) & 89.62 \\
    Reading Comprehension Specialist (Post-question) & 89.68 \\
    Reading Comprehension Specialist (Split) & 89.25 \\
    \midrule
    \multicolumn{2}{l}{\textit{RACE-m}} \\
    Reading Comprehension Specialist (Pre-question) & 93.66 \\
    Reading Comprehension Specialist (Post-question) & 93.80 \\
    Reading Comprehension Specialist (Split) & 93.87 \\
    \midrule
    \multicolumn{2}{l}{\textit{TriviaQA}} \\
    Data Scientist (Pre-question) & 84.10 \\
    Data Scientist (Post-question) & 82.80 \\
    Data Scientist (Split) & 84.50 \\
    \midrule
    \bottomrule
  \end{tabular}
} 
\end{table}

\begin{table}[!t]
  \caption{Performance of dynamically-selected roles for GPT-4.1-nano.}
  \label{dynamic_GPT41nano}
  \centering
 \resizebox{0.8\linewidth}{!}{ 
  \begin{tabular}{lc}
    \toprule
    \textbf{Dataset and Prompt Variant} & \textbf{Accuracy} \\
    \midrule
    \multicolumn{2}{l}{\textit{AQuA-RAT}} \\
    Mathematics Educator (Pre-question) & 69.69 \\
    Mathematics Educator (Post-question) & 66.14 \\
    Mathematics Educator (Split) & 70.87 \\
    \midrule
    \multicolumn{2}{l}{\textit{GSM8K}} \\
    Mathematics Educator (Pre-question) & 85.37 \\
    Mathematics Educator (Post-question) & 85.14 \\
    Mathematics Educator (Split) & 84.91 \\
    \midrule
    \multicolumn{2}{l}{\textit{OpenbookQA}} \\
    Science Educator (Pre-question) & 89.80 \\
    Science Educator (Post-question) & 90.00 \\
    Science Educator (Split) & 87.20 \\
    \midrule
    \multicolumn{2}{l}{\textit{StrategyQA}} \\
    Logic Expert (Pre-question) & 74.28 \\
    Logic Expert (Post-question) & 73.62 \\
    Logic Expert (Split) & 74.67 \\
    \midrule
    \multicolumn{2}{l}{\textit{RACE-h}} \\
    Educational Researcher (Pre-question) & 79.33 \\
    Educational Researcher (Post-question) & 82.53 \\
    Educational Researcher (Split) & 82.50 \\
    \midrule
    \multicolumn{2}{l}{\textit{RACE-m}} \\
    Educational Researcher (Pre-question) & 88.23 \\
    Educational Researcher (Post-question) & 89.35 \\
    Educational Researcher (Split) & 90.04 \\
    \midrule
    \multicolumn{2}{l}{\textit{TriviaQA}} \\
    Reading Comprehension Researcher (Pre-question) & 76.40 \\
    Reading Comprehension Researcher (Post-question) & 78.70 \\
    Reading Comprehension Researcher (Split) & 80.30 \\
    \midrule
    \bottomrule
  \end{tabular}
} 
\end{table}

\begin{table}[!t]
  \caption{Performance of dynamically-selected roles for GPT-4o.}
  \label{dynamic_GPT4o}
  \centering
 \resizebox{0.8\linewidth}{!}{ 
  \begin{tabular}{lc}
    \toprule
    \textbf{Dataset and Prompt Variant} & \textbf{Accuracy} \\
    \midrule
    \multicolumn{2}{l}{\textit{AQuA-RAT}} \\
    (Same as Static) & $\equiv$ \\
    \midrule
    \multicolumn{2}{l}{\textit{GSM8K}} \\
    (Same as Static) & $\equiv$ \\
    \midrule
    \multicolumn{2}{l}{\textit{OpenbookQA}} \\
    Cognitive Scientist (Pre-question) & 92.60 \\
    Cognitive Scientist (Post-question) & 89.40 \\
    Cognitive Scientist (Split) & 91.20 \\
    \midrule
    \multicolumn{2}{l}{\textit{StrategyQA}} \\
    Cognitive Scientist (Pre-question) & 82.05 \\
    Cognitive Scientist (Post-question) & 80.61 \\
    Cognitive Scientist (Split) & 82.27 \\
    \midrule
    \multicolumn{2}{l}{\textit{RACE-h}} \\
    English Teacher (Pre-question) & 89.54 \\
    English Teacher (Post-question) & 88.42 \\
    English Teacher (Split) & 89.57 \\
    \midrule
    \multicolumn{2}{l}{\textit{RACE-m}} \\
    English Teacher (Pre-question) & 94.22 \\
    English Teacher (Post-question) & 93.11 \\
    English Teacher (Split) & 94.01 \\
    \midrule
    \multicolumn{2}{l}{\textit{TriviaQA}} \\
    Trivia Expert (Pre-question) & 87.00 \\
    Trivia Expert (Post-question) & 87.00 \\
    Trivia Expert (Split) & 87.00 \\
    \midrule
    \bottomrule
  \end{tabular}
} 
\end{table}

\begin{table}[!t]
  \caption{Performance of dynamically-selected roles for GPT-4o-mini.}
  \label{dynamic_GPT4omini}
  \centering
 \resizebox{0.8\linewidth}{!}{ 
  \begin{tabular}{lc}
    \toprule
    \textbf{Dataset and Prompt Variant} & \textbf{Accuracy} \\
    \midrule
    \multicolumn{2}{l}{\textit{AQuA-RAT}} \\
    Mathematics Educator (Pre-question) & 72.44 \\
    Mathematics Educator (Post-question) & 73.62 \\
    Mathematics Educator (Split) & 70.47 \\
    \midrule
    \multicolumn{2}{l}{\textit{GSM8K}} \\
    (Same as Static) & $\equiv$ \\
    \midrule
    \multicolumn{2}{l}{\textit{OpenbookQA}} \\
    Research Scientist (Pre-question) & 83.60 \\
    Research Scientist (Post-question) & 73.40 \\
    Research Scientist (Split) & 75.20 \\
    \midrule
    \multicolumn{2}{l}{\textit{StrategyQA}} \\
    Data Scientist (Pre-question) & 78.91 \\
    Data Scientist (Post-question) & 78.60 \\
    Data Scientist (Split) & 79.34 \\
    \midrule
    \multicolumn{2}{l}{\textit{RACE-h}} \\
    Educational Assessment Specialist (Pre-question) & 86.54 \\
    Educational Assessment Specialist (Post-question) & 85.31 \\
    Educational Assessment Specialist (Split) & 85.16 \\
    \midrule
    \multicolumn{2}{l}{\textit{RACE-m}} \\
    Educational Assessment Specialist (Pre-question) & 91.85 \\
    Educational Assessment Specialist (Post-question) & 92.62 \\
    Educational Assessment Specialist (Split) & 92.62 \\
    \midrule
    \multicolumn{2}{l}{\textit{TriviaQA}} \\
    Data Scientist (Pre-question) & 79.20 \\
    Data Scientist (Post-question) & 79.50 \\
    Data Scientist (Split) & 79.90 \\
    \midrule
    \bottomrule
  \end{tabular}
} 
\end{table}

\begin{table}[!t]
  \caption{Performance of dynamically-selected roles for Phi4-14b.}
  \label{dynamic_Phi414b}
  \centering
 \resizebox{0.8\linewidth}{!}{ 
  \begin{tabular}{lc}
    \toprule
    \textbf{Dataset and Prompt Variant} & \textbf{Accuracy} \\
    \midrule
    \multicolumn{2}{l}{\textit{AQuA-RAT}} \\
    Mathematics Educator (Pre-question) & 53.15 \\
    Mathematics Educator (Post-question) & 65.35 \\
    Mathematics Educator (Split) & 54.33 \\
    \midrule
    \multicolumn{2}{l}{\textit{GSM8K}} \\
    Mathematics Educator (Pre-question) & 89.31 \\
    Mathematics Educator (Post-question) & 88.32 \\
    Mathematics Educator (Split) & 88.93 \\
    \midrule
    \multicolumn{2}{l}{\textit{OpenbookQA}} \\
    Cognitive Scientist (Pre-question) & 61.00 \\
    Cognitive Scientist (Post-question) & 61.20 \\
    Cognitive Scientist (Split) & 60.20 \\
    \midrule
    \multicolumn{2}{l}{\textit{StrategyQA}} \\
    Ai Research Scientist (Pre-question) & 76.94 \\
    Ai Research Scientist (Post-question) & 77.34 \\
    Ai Research Scientist (Split) & 76.90 \\
    \midrule
    \multicolumn{2}{l}{\textit{RACE-h}} \\
    Reading Specialist (Pre-question) & 83.73 \\
    Reading Specialist (Post-question) & 82.16 \\
    Reading Specialist (Split) & 82.19 \\
    \midrule
    \multicolumn{2}{l}{\textit{RACE-m}} \\
    Reading Specialist (Pre-question) & 90.67 \\
    Reading Specialist (Post-question) & 89.62 \\
    Reading Specialist (Split) & 89.90 \\
    \midrule
    \multicolumn{2}{l}{\textit{TriviaQA}} \\
    Knowledge Engineer (Pre-question) & 75.70 \\
    Knowledge Engineer (Post-question) & 75.50 \\
    Knowledge Engineer (Split) & 75.20 \\
    \midrule
    \bottomrule
  \end{tabular}
} 
\end{table}

\begin{table}[!t]
  \caption{Performance of dynamically-selected roles for Qwen2.5-14b.}
  \label{dynamic_Qwen2514b}
  \centering
 \resizebox{0.8\linewidth}{!}{ 
  \begin{tabular}{lc}
    \toprule
    \textbf{Dataset and Prompt Variant} & \textbf{Accuracy} \\
    \midrule
    \multicolumn{2}{l}{\textit{AQuA-RAT}} \\
    Mathematics Educator (Pre-question) & 60.24 \\
    Mathematics Educator (Post-question) & 59.84 \\
    Mathematics Educator (Split) & 50.39 \\
    \midrule
    \multicolumn{2}{l}{\textit{GSM8K}} \\
    Math Teacher (Pre-question) & 87.72 \\
    Math Teacher (Post-question) & 87.95 \\
    Math Teacher (Split) & 88.25 \\
    \midrule
    \multicolumn{2}{l}{\textit{OpenbookQA}} \\
    Knowledge Engineer (Pre-question) & 55.40 \\
    Knowledge Engineer (Post-question) & 53.60 \\
    Knowledge Engineer (Split) & 55.00 \\
    \midrule
    \multicolumn{2}{l}{\textit{StrategyQA}} \\
    Strategist (Pre-question) & 73.54 \\
    Strategist (Post-question) & 69.00 \\
    Strategist (Split) & 71.05 \\
    \midrule
    \multicolumn{2}{l}{\textit{RACE-h}} \\
    Educational Assessment Specialist (Pre-question) & 86.51 \\
    Educational Assessment Specialist (Post-question) & 86.82 \\
    Educational Assessment Specialist (Split) & 86.36 \\
    \midrule
    \multicolumn{2}{l}{\textit{RACE-m}} \\
    Educational Assessment Specialist (Pre-question) & 91.09 \\
    Educational Assessment Specialist (Post-question) & 90.60 \\
    Educational Assessment Specialist (Split) & 91.78 \\
    \midrule
    \multicolumn{2}{l}{\textit{TriviaQA}} \\
    Information Scientist (Pre-question) & 74.10 \\
    Information Scientist (Post-question) & 75.50 \\
    Information Scientist (Split) & 74.20 \\
    \midrule
    \bottomrule
  \end{tabular}
} 
\end{table}

\begin{table}[!t]
  \caption{Performance of dynamically-selected roles for Gemma3-12b.}
  \label{dynamic_Gemma312b}
  \centering
 \resizebox{0.8\linewidth}{!}{ 
  \begin{tabular}{lc}
    \toprule
    \textbf{Dataset and Prompt Variant} & \textbf{Accuracy} \\
    \midrule
    \multicolumn{2}{l}{\textit{AQuA-RAT}} \\
    Mathematics Educator (Pre-question) & 50.79 \\
    Mathematics Educator (Post-question) & 58.66 \\
    Mathematics Educator (Split) & 23.23 \\
    \midrule
    \multicolumn{2}{l}{\textit{GSM8K}} \\
    Mathematics Educator (Pre-question) & 83.09 \\
    Mathematics Educator (Post-question) & 88.02 \\
    Mathematics Educator (Split) & 65.05 \\
    \midrule
    \multicolumn{2}{l}{\textit{OpenbookQA}} \\
    Research Scientist (Pre-question) & 47.80 \\
    Research Scientist (Post-question) & 49.00 \\
    Research Scientist (Split) & 48.60 \\
    \midrule
    \multicolumn{2}{l}{\textit{StrategyQA}} \\
    Reasoning Analyst (Pre-question) & 74.76 \\
    Reasoning Analyst (Post-question) & 76.42 \\
    Reasoning Analyst (Split) & 75.11 \\
    \midrule
    \multicolumn{2}{l}{\textit{RACE-h}} \\
    Educator (Pre-question) & 84.02 \\
    Educator (Post-question) & 82.39 \\
    Educator (Split) & 83.22 \\
    \midrule
    \multicolumn{2}{l}{\textit{RACE-m}} \\
    Educator (Pre-question) & 88.79 \\
    Educator (Post-question) & 87.26 \\
    Educator (Split) & 86.35 \\
    \midrule
    \multicolumn{2}{l}{\textit{TriviaQA}} \\
    Natural Language Processing Scientist (Pre-question) & 70.80 \\
    Natural Language Processing Scientist (Post-question) & 70.50 \\
    Natural Language Processing Scientist (Split) & 71.30 \\
    \midrule
    \bottomrule
  \end{tabular}
} 
\end{table}

\begin{table}[!t]
  \caption{Performance of dynamically-selected roles for Mistral-nemo-12b.}
  \label{dynamic_Mistralnemo12b}
  \centering
 \resizebox{0.8\linewidth}{!}{ 
  \begin{tabular}{lc}
    \toprule
    \textbf{Dataset and Prompt Variant} & \textbf{Accuracy} \\
    \midrule
    \multicolumn{2}{l}{\textit{AQuA-RAT}} \\
    Mathematical Linguist (Pre-question) & 55.91 \\
    Mathematical Linguist (Post-question) & 59.84 \\
    Mathematical Linguist (Split) & 56.69 \\
    \midrule
    \multicolumn{2}{l}{\textit{GSM8K}} \\
    (Same as Static) & $\equiv$ \\
    \midrule
    \multicolumn{2}{l}{\textit{OpenbookQA}} \\
    Expert In Natural Language Processing (Pre-question) & 77.40 \\
    Expert In Natural Language Processing (Post-question) & 72.80 \\
    Expert In Natural Language Processing (Split) & 72.20 \\
    \midrule
    \multicolumn{2}{l}{\textit{StrategyQA}} \\
    Data Scientist (Pre-question) & 67.99 \\
    Data Scientist (Post-question) & 69.48 \\
    Data Scientist (Split) & 71.40 \\
    \midrule
    \multicolumn{2}{l}{\textit{RACE-h}} \\
    Educational Psychologist (Pre-question) & 45.63 \\
    Educational Psychologist (Post-question) & 56.20 \\
    Educational Psychologist (Split) & 58.20 \\
    \midrule
    \multicolumn{2}{l}{\textit{RACE-m}} \\
    Educational Psychologist (Pre-question) & 65.39 \\
    Educational Psychologist (Post-question) & 69.99 \\
    Educational Psychologist (Split) & 69.36 \\
    \midrule
    \multicolumn{2}{l}{\textit{TriviaQA}} \\
    Information Retrieval Specialist (Pre-question) & 74.60 \\
    Information Retrieval Specialist (Post-question) & 73.50 \\
    Information Retrieval Specialist (Split) & 73.20 \\
    \midrule
    \bottomrule
  \end{tabular}
} 
\end{table}

\begin{table}[!t]
  \caption{Performance of dynamically-selected roles for Gemma2-9b.}
  \label{dynamic_Gemma29b}
  \centering
 \resizebox{0.8\linewidth}{!}{ 
  \begin{tabular}{lc}
    \toprule
    \textbf{Dataset and Prompt Variant} & \textbf{Accuracy} \\
    \midrule
    \multicolumn{2}{l}{\textit{AQuA-RAT}} \\
    Mathematics Educator (Pre-question) & 51.57 \\
    Mathematics Educator (Post-question) & 50.39 \\
    Mathematics Educator (Split) & 49.61 \\
    \midrule
    \multicolumn{2}{l}{\textit{GSM8K}} \\
    Educational Psychologist (Pre-question) & 81.96 \\
    Educational Psychologist (Post-question) & 77.03 \\
    Educational Psychologist (Split) & 78.24 \\
    \midrule
    \multicolumn{2}{l}{\textit{OpenbookQA}} \\
    Research Scientist (Pre-question) & 33.60 \\
    Research Scientist (Post-question) & 32.40 \\
    Research Scientist (Split) & 32.60 \\
    \midrule
    \multicolumn{2}{l}{\textit{StrategyQA}} \\
    Research Scientist (Pre-question) & 67.12 \\
    Research Scientist (Post-question) & 65.28 \\
    Research Scientist (Split) & 66.07 \\
    \midrule
    \multicolumn{2}{l}{\textit{RACE-h}} \\
    Educator (Pre-question) & 77.79 \\
    Educator (Post-question) & 81.42 \\
    Educator (Split) & 81.07 \\
    \midrule
    \multicolumn{2}{l}{\textit{RACE-m}} \\
    Educator (Pre-question) & 88.86 \\
    Educator (Post-question) & 88.02 \\
    Educator (Split) & 88.72 \\
    \midrule
    \multicolumn{2}{l}{\textit{TriviaQA}} \\
    Research Scientist (Pre-question) & 66.50 \\
    Research Scientist (Post-question) & 63.20 \\
    Research Scientist (Split) & 62.80 \\
    \midrule
    \bottomrule
  \end{tabular}
} 
\end{table}

\begin{table}[!t]
  \caption{Performance of dynamically-selected roles for Granite3.2-8b.}
  \label{dynamic_Granite328b}
  \centering
 \resizebox{0.8\linewidth}{!}{ 
  \begin{tabular}{lc}
    \toprule
    \textbf{Dataset and Prompt Variant} & \textbf{Accuracy} \\
    \midrule
    \multicolumn{2}{l}{\textit{AQuA-RAT}} \\
    Mathematics Educator (Pre-question) & 58.27 \\
    Mathematics Educator (Post-question) & 59.06 \\
    Mathematics Educator (Split) & 59.45 \\
    \midrule
    \multicolumn{2}{l}{\textit{GSM8K}} \\
    Mathematics Educator (Pre-question) & 77.10 \\
    Mathematics Educator (Post-question) & 77.03 \\
    Mathematics Educator (Split) & 76.35 \\
    \midrule
    \multicolumn{2}{l}{\textit{OpenbookQA}} \\
    Natural Language Processing Specialist (Pre-question) & 50.20 \\
    Natural Language Processing Specialist (Post-question) & 57.80 \\
    Natural Language Processing Specialist (Split) & 62.80 \\
    \midrule
    \multicolumn{2}{l}{\textit{StrategyQA}} \\
    Data Scientist (Pre-question) & 63.80 \\
    Data Scientist (Post-question) & 61.53 \\
    Data Scientist (Split) & 65.15 \\
    \midrule
    \multicolumn{2}{l}{\textit{RACE-h}} \\
    Curriculum Developer (Pre-question) & 65.07 \\
    Curriculum Developer (Post-question) & 63.69 \\
    Curriculum Developer (Split) & 63.41 \\
    \midrule
    \multicolumn{2}{l}{\textit{RACE-m}} \\
    Curriculum Developer (Pre-question) & 72.35 \\
    Curriculum Developer (Post-question) & 73.47 \\
    Curriculum Developer (Split) & 72.77 \\
    \midrule
    \multicolumn{2}{l}{\textit{TriviaQA}} \\
    Natural Language Processing Specialist (Pre-question) & 69.70 \\
    Natural Language Processing Specialist (Post-question) & 69.80 \\
    Natural Language Processing Specialist (Split) & 69.80 \\
    \midrule
    \bottomrule
  \end{tabular}
} 
\end{table}

\begin{table}[!t]
  \caption{Performance of dynamically-selected roles for Granite3.3-8b.}
  \label{dynamic_Granite338b}
  \centering
 \resizebox{0.8\linewidth}{!}{ 
  \begin{tabular}{lc}
    \toprule
    \textbf{Dataset and Prompt Variant} & \textbf{Accuracy} \\
    \midrule
    \multicolumn{2}{l}{\textit{AQuA-RAT}} \\
    Curriculum Developer (Pre-question) & 59.84 \\
    Curriculum Developer (Post-question) & 55.91 \\
    Curriculum Developer (Split) & 60.63 \\
    \midrule
    \multicolumn{2}{l}{\textit{GSM8K}} \\
    Curriculum Developer (Pre-question) & 77.33 \\
    Curriculum Developer (Post-question) & 77.79 \\
    Curriculum Developer (Split) & 78.24 \\
    \midrule
    \multicolumn{2}{l}{\textit{OpenbookQA}} \\
    Knowledge Engineer (Pre-question) & 71.00 \\
    Knowledge Engineer (Post-question) & 74.40 \\
    Knowledge Engineer (Split) & 75.20 \\
    \midrule
    \multicolumn{2}{l}{\textit{StrategyQA}} \\
    Strategy Analyst (Pre-question) & 64.32 \\
    Strategy Analyst (Post-question) & 67.69 \\
    Strategy Analyst (Split) & 64.80 \\
    \midrule
    \multicolumn{2}{l}{\textit{RACE-h}} \\
    Curriculum Developer (Pre-question) & 69.47 \\
    Curriculum Developer (Post-question) & 69.81 \\
    Curriculum Developer (Split) & 72.44 \\
    \midrule
    \multicolumn{2}{l}{\textit{RACE-m}} \\
    Curriculum Developer (Pre-question) & 79.60 \\
    Curriculum Developer (Post-question) & 77.51 \\
    Curriculum Developer (Split) & 80.50 \\
    \midrule
    \multicolumn{2}{l}{\textit{TriviaQA}} \\
    Nlp Researcher (Pre-question) & 66.20 \\
    Nlp Researcher (Post-question) & 66.70 \\
    Nlp Researcher (Split) & 64.30 \\
    \midrule
    \bottomrule
  \end{tabular}
} 
\end{table}

\begin{table}[!t]
  \caption{Performance of dynamically-selected roles for Llama3.1-8b.}
  \label{dynamic_Llama318b}
  \centering
 \resizebox{0.8\linewidth}{!}{ 
  \begin{tabular}{lc}
    \toprule
    \textbf{Dataset and Prompt Variant} & \textbf{Accuracy} \\
    \midrule
    \multicolumn{2}{l}{\textit{AQuA-RAT}} \\
    (Same as Static) & $\equiv$ \\
    \midrule
    \multicolumn{2}{l}{\textit{GSM8K}} \\
    Math Educator (Pre-question) & 80.36 \\
    Math Educator (Post-question) & 79.61 \\
    Math Educator (Split) & 79.76 \\
    \midrule
    \multicolumn{2}{l}{\textit{OpenbookQA}} \\
    Research Scientist (Pre-question) & 48.80 \\
    Research Scientist (Post-question) & 35.40 \\
    Research Scientist (Split) & 55.00 \\
    \midrule
    \multicolumn{2}{l}{\textit{StrategyQA}} \\
    Lawyer (Pre-question) & 69.69 \\
    Lawyer (Post-question) & 66.38 \\
    Lawyer (Split) & 68.60 \\
    \midrule
    \multicolumn{2}{l}{\textit{RACE-h}} \\
    English Teacher (Pre-question) & 77.84 \\
    English Teacher (Post-question) & 76.19 \\
    English Teacher (Split) & 78.22 \\
    \midrule
    \multicolumn{2}{l}{\textit{RACE-m}} \\
    English Teacher (Pre-question) & 83.70 \\
    English Teacher (Post-question) & 83.29 \\
    English Teacher (Split) & 82.94 \\
    \midrule
    \multicolumn{2}{l}{\textit{TriviaQA}} \\
    Information Architect (Pre-question) & 74.40 \\
    Information Architect (Post-question) & 75.40 \\
    Information Architect (Split) & 73.80 \\
    \midrule
    \bottomrule
  \end{tabular}
} 
\end{table}

\begin{table}[!t]
  \caption{Performance of dynamically-selected roles for Mistral-7b.}
  \label{dynamic_Mistral7b}
  \centering
 \resizebox{0.8\linewidth}{!}{ 
  \begin{tabular}{lc}
    \toprule
    \textbf{Dataset and Prompt Variant} & \textbf{Accuracy} \\
    \midrule
    \multicolumn{2}{l}{\textit{AQuA-RAT}} \\
    (Same as Static) & $\equiv$ \\
    \midrule
    \multicolumn{2}{l}{\textit{GSM8K}} \\
    Data Scientist (Pre-question) & 38.36 \\
    Data Scientist (Post-question) & 41.47 \\
    Data Scientist (Split) & 42.53 \\
    \midrule
    \multicolumn{2}{l}{\textit{OpenbookQA}} \\
    Scientist (Pre-question) & 39.60 \\
    Scientist (Post-question) & 38.40 \\
    Scientist (Split) & 41.60 \\
    \midrule
    \multicolumn{2}{l}{\textit{StrategyQA}} \\
    Data Scientist (Pre-question) & 38.47 \\
    Data Scientist (Post-question) & 36.72 \\
    Data Scientist (Split) & 38.43 \\
    \midrule
    \multicolumn{2}{l}{\textit{RACE-h}} \\
    (Same as Static) & $\equiv$ \\
    \midrule
    \multicolumn{2}{l}{\textit{RACE-m}} \\
    (Same as Static) & $\equiv$ \\
    \midrule
    \multicolumn{2}{l}{\textit{TriviaQA}} \\
    Data Scientist (Pre-question) & 31.20 \\
    Data Scientist (Post-question) & 46.60 \\
    Data Scientist (Split) & 35.80 \\
    \midrule
    \bottomrule
  \end{tabular}
} 
\end{table}

\begin{table}[!t]
  \caption{Performance of dynamically-selected roles for Qwen2.5-7b.}
  \label{dynamic_Qwen257b}
  \centering
 \resizebox{0.8\linewidth}{!}{ 
  \begin{tabular}{lc}
    \toprule
    \textbf{Dataset and Prompt Variant} & \textbf{Accuracy} \\
    \midrule
    \multicolumn{2}{l}{\textit{AQuA-RAT}} \\
    (Same as Static) & $\equiv$ \\
    \midrule
    \multicolumn{2}{l}{\textit{GSM8K}} \\
    Math Teacher (Pre-question) & 80.44 \\
    Math Teacher (Post-question) & 84.91 \\
    Math Teacher (Split) & 85.29 \\
    \midrule
    \multicolumn{2}{l}{\textit{OpenbookQA}} \\
    Knowledge Engineer (Pre-question) & 67.00 \\
    Knowledge Engineer (Post-question) & 46.00 \\
    Knowledge Engineer (Split) & 47.80 \\
    \midrule
    \multicolumn{2}{l}{\textit{StrategyQA}} \\
    Information Retrieval Specialist (Pre-question) & 62.27 \\
    Information Retrieval Specialist (Post-question) & 64.98 \\
    Information Retrieval Specialist (Split) & 66.42 \\
    \midrule
    \multicolumn{2}{l}{\textit{RACE-h}} \\
    Natural Language Processing Specialist (Pre-question) & 80.99 \\
    Natural Language Processing Specialist (Post-question) & 80.96 \\
    Natural Language Processing Specialist (Split) & 82.02 \\
    \midrule
    \multicolumn{2}{l}{\textit{RACE-m}} \\
    Natural Language Processing Specialist (Pre-question) & 87.74 \\
    Natural Language Processing Specialist (Post-question) & 87.19 \\
    Natural Language Processing Specialist (Split) & 88.79 \\
    \midrule
    \multicolumn{2}{l}{\textit{TriviaQA}} \\
    Information Retrieval Specialist (Pre-question) & 63.40 \\
    Information Retrieval Specialist (Post-question) & 65.10 \\
    Information Retrieval Specialist (Split) & 64.90 \\
    \midrule
    \bottomrule
  \end{tabular}
} 
\end{table}

\begin{table}[!t]
  \caption{Performance of dynamically-selected roles for Llama3.2-3b.}
  \label{dynamic_Llama323b}
  \centering
 \resizebox{0.8\linewidth}{!}{ 
  %
} 
\end{table}

\begin{table}[!t]
  \caption{Performance of dynamically-selected roles for Qwen2.5-3b.}
  \label{dynamic_Qwen253b}
  \centering
 \resizebox{0.8\linewidth}{!}{ 
  %
} 
\end{table}

\begin{table}[!t]
  \caption{Performance of dynamically-selected roles for Gemma2-2b.}
  \label{dynamic_Gemma22b}
  \centering
 \resizebox{0.8\linewidth}{!}{ 
  %
} 
\end{table}

\begin{table}[!t]
  \caption{Performance of dynamically-selected roles for Granite3.2-2b.}
  \label{dynamic_Granite322b}
  \centering
 \resizebox{0.8\linewidth}{!}{ 
  %
} 
\end{table}

\end{document}